\pdfoutput=1

\documentclass[11pt]{article}

\usepackage{acl}

\usepackage{times}
\usepackage{latexsym}

\usepackage[T1]{fontenc}

\usepackage[utf8]{inputenc}

\usepackage{microtype}

\usepackage{bm}
\usepackage{amsmath}
\usepackage{amsfonts}
\usepackage{amssymb}
\usepackage{graphicx}
\usepackage{booktabs}
\usepackage{multirow}
\usepackage{makecell}
\usepackage{enumitem}
\usepackage{algorithm}
\usepackage{algorithmic}
\usepackage{xspace}

\usepackage{color} 
\usepackage{url}
\usepackage{fontawesome}

\newenvironment{des}{ 
     \parskip 0cm \begin{list}{}{\parsep 0cm \itemsep 0cm \topsep 0cm}}{
       \end{list}} 

\newcommand{\rw}[1]{\textcolor[rgb]{0.00,0.00,0.00}{#1}}

\newcommand{\modelname}{\textsc{MetGen}\xspace}
\newcommand{\with}{+}

\newcommand{\EntialmentWriter}{EntialmentWriter\xspace}
\newcommand{\EntailmentWriterIter}{EntialmentWriter-Iter\xspace}

\newcommand{\Oursep}{{\modelname-separated}\xspace}
\newcommand{\Ourpre}{{\modelname-prefixed}\xspace}

\title{\textsc{MetGen}: A Module-Based Entailment Tree Generation Framework for Answer Explanation}

\author{Ruixin Hong\textsuperscript{1}, Hongming Zhang\textsuperscript{2}, Xintong Yu\textsuperscript{1}, Changshui Zhang\textsuperscript{1} \\
\textsuperscript{1}Institute for Artificial Intelligence, Tsinghua University (THUAI); \\
\textsuperscript{1}Beijing National Research Center for Information Science and Technology (BNRist); \\
\textsuperscript{1}Department of Automation, Tsinghua University, Beijing, P.R.China \\
\textsuperscript{2}Tencent AI Lab, Seattle \\
\texttt{\{hrx20, yuxt16\}@mails.tsinghua.edu.cn,} \\
\texttt{hongmzhang@tencent.com,}
\texttt{zcs@mail.tsinghua.edu.cn,}
}

\begin{document}
\maketitle
\begin{abstract}
Knowing the reasoning chains from knowledge to the predicted answers can help construct an explainable question answering (QA) system.
Advances on QA explanation propose to explain the answers with entailment trees composed of multiple entailment steps.
While current work proposes to generate entailment trees with end-to-end generative models, the steps in the generated trees are not constrained and could be unreliable.
In this paper, we propose \modelname, a Module-based Entailment Tree GENeration framework that has multiple modules and a reasoning controller.
Given a question and several supporting knowledge, \modelname can iteratively generate the entailment tree by conducting single-step entailment with separate modules and selecting the reasoning flow with the controller. 
As each module is guided to perform a specific type of entailment reasoning, the steps generated by \modelname are more reliable and valid.
Experiment results on the standard benchmark show that \modelname can outperform previous state-of-the-art models with only 9\% of the parameters.

\end{abstract}

\section{Introduction}

{\setlength{\abovecaptionskip}{2mm}
\begin{figure}[t!]
\centering
\includegraphics[width=\columnwidth]{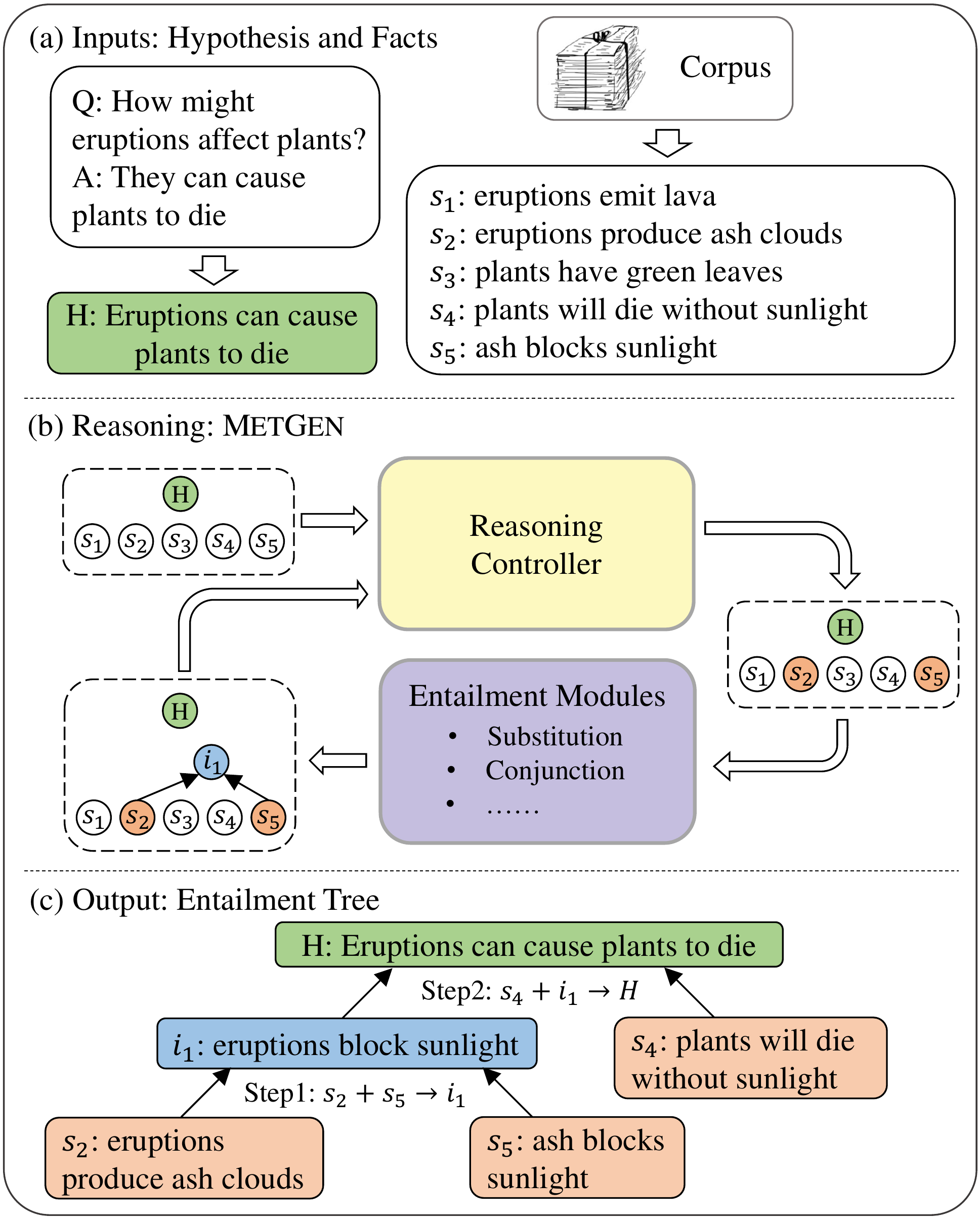}
\caption{
Given facts related to the question+answer, \modelname iteratively generates an entailment tree that contains the hypothesis (green), used facts (orange), and intermediate conclusions (blue) with several separate entailment modules and a reasoning controller.
}
\label{fig:task_intro}
\end{figure}
}

Explanation is recognized as a key factor toward responsible AI systems~\cite{DBLP:journals/inffus/ArrietaRSBTBGGM20}.
In the context of question answering (QA), providing an explanation of the predicted answers can help improve the understandability, debuggability, and trustworthiness of QA models.
Great efforts have been devoted to revealing how the models predict the answers and give explanations in various forms, including showing an attention map over passages~\cite{DBLP:conf/iclr/SeoKFH17}, 
giving a snippet of textual evidence~\cite{DBLP:conf/acl/DeYoungJRLXSW20},
and selecting answer-supporting sentences~\cite{DBLP:conf/lrec/XieTMWMJ20,DBLP:conf/textgraphs/JansenU19}.
Among all explanation forms, the \emph{entailment trees}~\cite{DBLP:conf/emnlp/DalviJTXSPC21} provide the most detailed and informative explanation by exposing the \emph{chains of reasoning} from the knowledge to the predictions. 
As shown in Figure~\ref{fig:task_intro}(a) and (c), given a hypothesis (summarizing a question+answer pair) and supporting facts (retrieved from a corpus), the goal is to generate an entailment tree where each non-leaf node is an entailment of its children.
Providing a valid entailment tree would help users to understand how the hypothesis is proved, obtain novel intermediate conclusions from the basic knowledge, and gain detailed information to support decision making.

To generate the entailment trees, \citet{DBLP:conf/emnlp/DalviJTXSPC21} propose EntailmentWriter, an end-to-end sequence-to-sequence generative model, trained by maximizing the generation likelihood of the linearized gold trees.
However, they do not have an explicit strategy to constrain the validity of every single step and the tree structure.
Thus, the steps are not guaranteed to satisfy the reasoning rules and could be incorrect and unreliable.
For example, the step conclusion may not be entailed by the input premises or simply repeat one of the input premises~\cite{DBLP:conf/emnlp/DalviJTXSPC21}.
Furthermore, although their outputs are trees that can indicate the reasoning chains, the mapping mechanisms from the inputs to the trees remain implicit and invisible.

To tackle the above problems, we propose \modelname, a module-based framework to generate entailment trees in a more explicit approach and constrain the entailment steps with reasoning rules.
As shown in Figure~\ref{fig:task_intro}(b),
given the target hypothesis and known facts, \modelname first uses the reasoning controller to select some steps that can help get closer to the hypothesis.
Subsequently, \modelname executes the selected steps with single-step entailment modules and adds the generated intermediate facts into the known facts for the next round of reasoning.
Through this iterative approach, \modelname proves the hypothesis step by step and generates the overall entailment tree.

Each module in \modelname is a generative model that can perform a specific type of entailment reasoning (e.g., making a substitution inference).
To guide the modules to generate correct and sound conclusions, we train the modules with well-formed synthetic data containing the corresponding logical regularities of the reasoning types~\cite{DBLP:conf/emnlp/BostromZCD21}.
Inspired by the forward chaining and backward chaining algorithms in logic programming~\cite{chein2008graph}, we adopt both deductive and abductive modules to execute forward and backward reasoning steps, respectively.

Experiments on the standard benchmark EntailmentBank~\cite{DBLP:conf/emnlp/DalviJTXSPC21} show that \modelname can outperform the previous best model with 9.0\% of the model parameters.
Manual evaluation results demonstrate that \modelname can generate more reliable steps.
Further experiments under the data-scarce setting and cross-dataset setting (on eQASC and eOBQA~\cite{DBLP:conf/emnlp/JhamtaniC20}) show that \modelname is more data-efficient and has better generalization capability compared with the baselines.

{
\setlength{\abovecaptionskip}{2mm}
\begin{figure*}[th!]
\centering
\includegraphics[width=\textwidth]{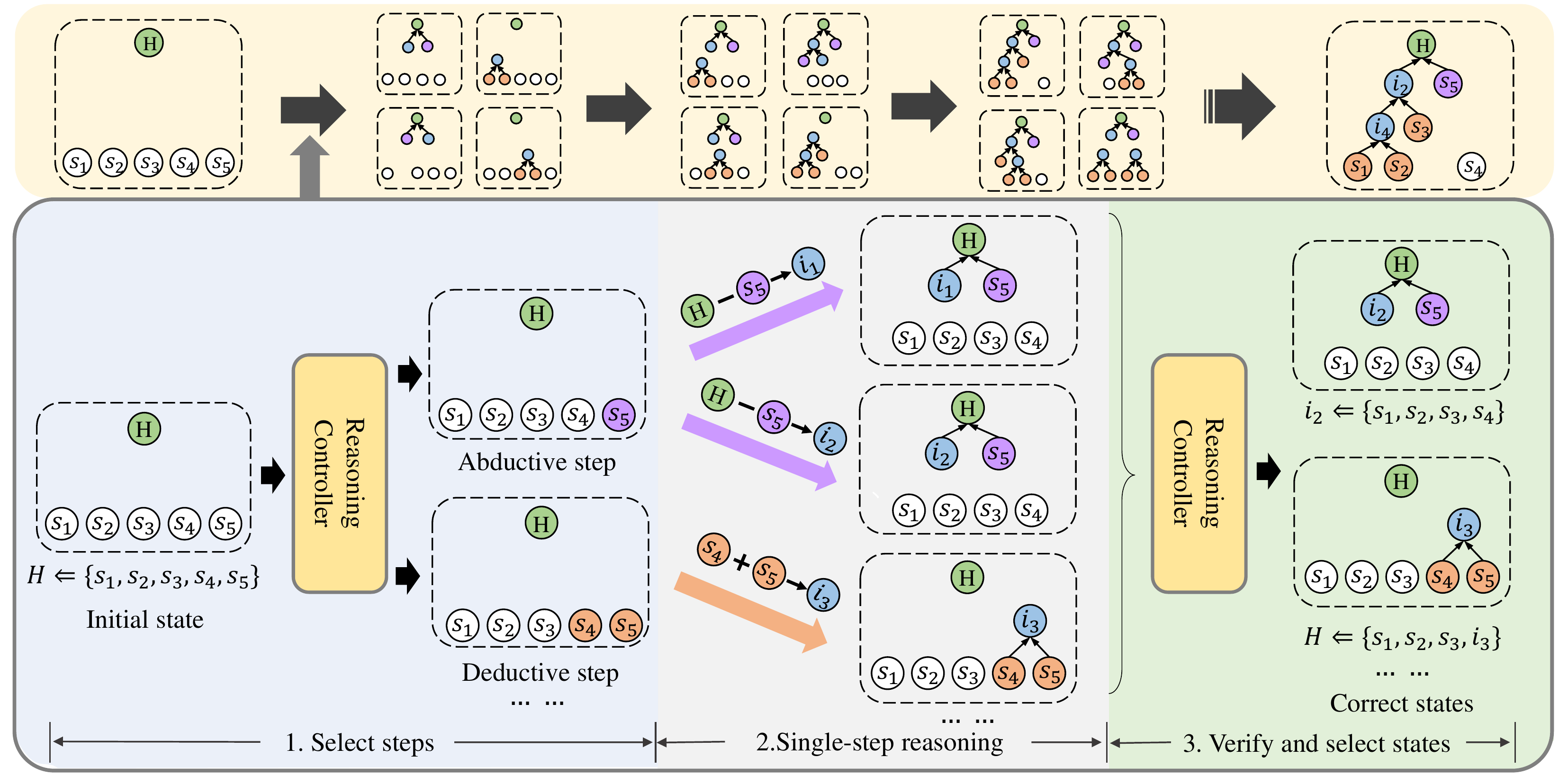}
\caption{Reasoning process of \modelname framework.
The goal is to prove the hypothesis with the given facts through reasoning iterations (the upper part). 
In the first reasoning iteration (the lower part), the initial state is denoted as $H \Leftarrow \{s_1,s_2,s_3,s_4,s_5\}$.
First, the controller selects promising steps, such as the backward abductive step $H-s_5$ and the forward deductive one $s_4+s_5$.
Then, single-step entailment modules perform the reasoning steps and generate novel intermediate facts including $i_1,i_2,i_3$.
After that, the controller verifies that 
the states $i_2 \Leftarrow \{s_1,s_2,s_3,s_4\}$ and $H \Leftarrow \{s_1,s_2,s_3,i_1\}$ are closer to the completion of reasoning and thus selects them for the next reasoning iteration.
}
\vspace{-5mm}
\label{fig:search}
\end{figure*}
}

\section{Related Works}

\textbf{Explainability in Question Answering.}
Recent works have explored the explainability of QA in various forms~\cite{DBLP:conf/iclr/SeoKFH17,DBLP:conf/emnlp/YeHBR20,DBLP:conf/emnlp/DalviJTXSPC21, DBLP:journals/tacl/LammPAACSC21, wiegreffe2021teach, DBLP:journals/corr/abs-2010-00389, DBLP:journals/corr/abs-2112-07772}.
One way is to retrieve multiple supporting facts related to the question or answer~\cite{DBLP:conf/lrec/XieTMWMJ20,DBLP:conf/textgraphs/JansenU19,DBLP:conf/emnlp/JhamtaniC20,DBLP:conf/acl/InoueSI20, DBLP:conf/emnlp/YadavBS19,DBLP:conf/acl/YadavBS20,DBLP:conf/eacl/ValentinoTF21,DBLP:conf/coling/CartuyvelsSM20,DBLP:conf/acl/ZhangZS20}.
These ``rationales''~\cite{DBLP:conf/acl/DeYoungJRLXSW20} provide insights about \emph{what} are used by the model to inform its predictions, but do not show \emph{how} the facts are combined to generate novel intermediate conclusions.
Some other works explain QA systems in a generative way, including generating explanation sentences that directly link a question to an answer~\cite{DBLP:conf/nips/CamburuRLB18,DBLP:conf/acl/RajaniMXS19} and thus expose the relevant knowledge used by models~\cite{DBLP:journals/corr/abs-2004-05569,DBLP:conf/emnlp/ShwartzWBBC20}.
However, as these models generate explanations in a free form, the generated facts may not be necessarily sound~\cite{DBLP:conf/emnlp/BostromZCD21}.
Recently, \citet{DBLP:conf/emnlp/BostromZCD21} propose ParaPattern, an automated pipeline for building two kinds of single-step deductions.
Different from the above work, our method generates the explanations in a multi-step tree structure~\cite{DBLP:conf/emnlp/DalviJTXSPC21}, showing \emph{what} and \emph{how} facts are combined to draw novel intermediate conclusions and reach the final answer.
The intermediate conclusions are generated by deductive and abductive entailment modules that are constrained to perform specific types of reasoning.

\noindent\textbf{Multi-Hop Proof Generation.}
Recently, several works propose to use the transformers for multi-hop logical reasoning and generate reliable formal proofs~\cite{DBLP:conf/ijcai/ClarkTR20,DBLP:conf/nips/TalmorTCGB20,DBLP:conf/emnlp/SahaGSB20,DBLP:conf/naacl/SahaYB21,DBLP:conf/acl/TafjordDC21}.
However, they mainly focus on synthetic sentences, which have low linguistic variation and struggle to represent the flexible sentences in real QA scenarios.

\noindent\textbf{Neural Module Networks.}
Decomposing the reasoning process into several pre-defined operations overlaps with the idea of neural module networks~\cite{DBLP:conf/cvpr/AndreasRDK16,DBLP:conf/iccv/HuARDS17,DBLP:conf/emnlp/GuptaL18,DBLP:conf/iclr/GuptaLR0020,DBLP:conf/acl/JiangJCB19}.
They typically assume that the question could be parsed into an executable program, i.e., the question \emph{explicitly} describes the process to arrive at the answer.
In our work, we tackle the questions/hypotheses that do not trivially describe the reasoning process and could be more challenging.

\section{Task Definition}
As shown in Figure~\ref{fig:task_intro}, the inputs are a hypothesis $H$ and some fact sentences $S = \{s_1,s_2,\dots,s_n \}$ (including both relevant and irrelevant ones) expressing knowledge.
$H$ is a declarative sentence derived from a question+answer pair and can be proved by the knowledge in $S$.
The desired output is a valid entailment tree $T$ with the root node being $H$, the leaves being facts selected from $S$, and the intermediate nodes being novel intermediate facts (e.g., $i_1,i_2$).
$T$ is considered valid if each non-leaf node is a valid entailment (a conclusion that ``a person would typically infer''~ \cite{DBLP:series/synthesis/2013Dagan}) of its immediate children.
We denote the annotated gold tree as $T_{gold}$ and its leaf facts as $S_{gold}$.
Following~\citet{DBLP:conf/emnlp/DalviJTXSPC21}, we consider three increasingly difficult tasks with different $S$:

{\spaceskip=0.12em\relax 
\begin{des}
\item[{\bf Task1 (no-distractor):}] $S = $ $S_{gold}$,
\item[{\bf Task2 (distractor):}] $S = $ $S_{gold}$ + 15-20 distractors,
\item[{\bf Task3 (full-corpus):}] $S = $ a corpus $C$.
\end{des}
}

\section{\modelname}

Figure~\ref{fig:search} illustrates the reasoning process of \modelname.
We reason one step at a time and iteratively generate the entailment trees.
In each iteration, given a reasoning state (e.g., the initial state $R_0: H \Leftarrow S$, where we aim to prove $H$ using $S$), the reasoning controller selects promising steps, including forward deductive steps and backward abductive ones.
We then use the corresponding modules to perform single-step entailment on the selected steps and generate novel intermediate facts.
Finally, we use the controller to verify the generated facts and select the correct states to perform further reasoning.
We introduce details about the module design, reasoning controller, and reasoning algorithm in Sec~\ref{sec:model_module}, ~\ref{sec:model_control}, and ~\ref{sec:model_alg}, respectively.

{\setlength{\abovecaptionskip}{2mm}
\begin{table*}[t!]
\centering
\includegraphics[width=\textwidth]{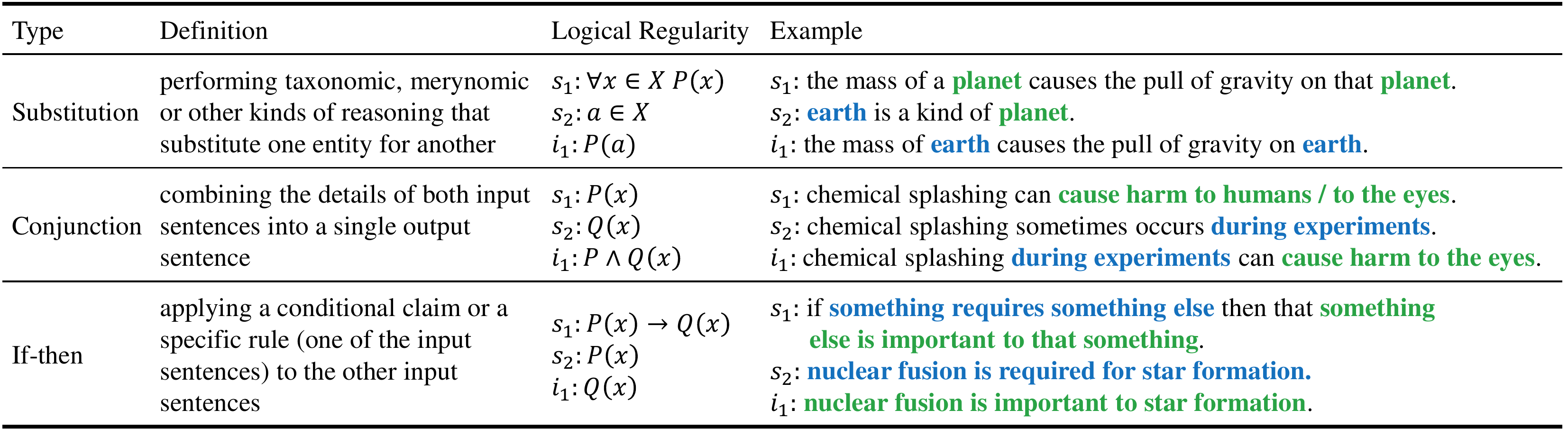}
\caption{
The used reasoning types.
Here, $s_1$ and $s_2$ denote input premises for deductive modules, while $i_1$ denotes the entailed conclusion.
For logical regularity, $P(x)$ means that the predicate $P$ is true for the entity $x$.
}
\vspace{-5mm}
\label{tab:module_examle}
\end{table*}
}

\subsection{Single-step Entailment Modules}
\label{sec:model_module}

\subsubsection{Module Definition}
We propose to divide the single-step entailment reasoning ability into a set of well-defined basic logical operations.
Such a design could help improve the generalization capability~\cite{DBLP:conf/emnlp/BostromZCD21,rudin2019stop}.
As shown in Table~\ref{tab:module_examle}, we adopt three common reasoning types, covering over 90\% of the steps in EntailmentBank according to the analysis by~\citet{DBLP:conf/emnlp/DalviJTXSPC21}.
Note that the entailment module types could be adjusted according to the specific tasks or domains, which allows our method to be flexibly applied to other problems.

We adopt both the deductive and abductive versions of the reasoning types.
Take a gold step $s_1\with s_2 \rightarrow i_1$ as an example.
Deduction is the process of reasoning from the premises to reach a logical conclusion.
A deductive module takes the two premises $s_1$ and $s_2$ as inputs and outputs a conclusion $\hat{i}_1$ according to its reasoning types (denoted as $s_1\with s_2 \rightarrow \hat{i}_1$).
Abduction is to find the best explanation given complete/incomplete observations~\cite{harman1965inference}.
In the context of the entailment steps, given a conclusion $i_1$ and a premise fact $s_2$ as observations, the abductive module yields a plausible premise $\hat{s}_1$ (denoted as $i_1 - s_2 \rightarrow \hat{s}_1$), where the generated premise $\hat{s}_1$ and the observed premise $s_2$ would most likely infer the conclusion $i_1$.
Although the steps in the EntailmentBank may have more than two premises, we only consider the case of two premises.
The reason is that the $n$-premise step ($n > 2$) could be further decomposed into several valid 2-premise steps~\cite{DBLP:conf/emnlp/DalviJTXSPC21} 
(See Appendix Figure~\ref{fig:case1} for a specific example).

\subsubsection{Module Training}
Training the entailment modules with data that contains the corresponding logical regularities would guide them to perform correct inferences and ensure soundness~\cite{DBLP:conf/emnlp/BostromZCD21}.
We first train the modules with synthetic sentences to learn the logical transformations and then further fine-tune them with the end task.

We follow ParePattern~\cite{DBLP:conf/emnlp/BostromZCD21}, a pipeline based on syntactic retrieval, rule-based example construction, and automatic paraphrasing, to collect synthetic sentences from Wikipedia.
Since~\citet{DBLP:conf/emnlp/BostromZCD21} only consider the substitution and contraposition deductions, we extend the method to conjunction and if-then deductions by designing the specific syntactic templates and construction rules (See Appendix~\ref{sec:parapattern_data}).
In addition, we also considered the abductive form of these modules.
We then fine-tune the modules with corresponding steps in EntailmentBank to adapt the modules to the science domain.
Since the original steps in EntailmentBank are not annotated with reasoning types, we manually label 400 steps of the training split and train a classifier with these steps.
The remaining steps are labeled with the pseudo labels predicted by the classifier.
We \emph{freeze} the parameters of modules once the training is complete.

\subsection{Reasoning Controller}
\label{sec:model_control}
In addition to single-step reasoning modules, we need to search for the correct path to reach the target hypothesis.
The entire reasoning search space would grow rapidly as the number of input facts increases and there would also be complex branching in the trees.
We introduce a reasoning controller to filter out incorrect facts, steps, and states to reduce the search space and complete the reasoning accurately and efficiently.

Figure~\ref{fig:search} shows how the controller is used in each reasoning iteration.
At the beginning of the iteration, the controller scores all possible \textit{steps} and selects the most promising ones for single-step entailment.
After the entailment modules generate intermediate facts, the controller estimates which \textit{state} with a generated fact gets closer to the completion of reasoning and selects the best states for the next iteration.
Besides the usage within each iteration, the controller also rates all \textit{facts} at the start of the whole reasoning process and keeps only the relevant facts for the initial state when fact distractors exist.

\subsubsection{Controller Model}

The controller model scores steps, facts, and states based on a transformer, and its structure is shown in Figure~\ref{fig:controller}.

{\setlength{\abovecaptionskip}{1mm}
\begin{figure}[t!]
\centering
\includegraphics[width=\columnwidth]{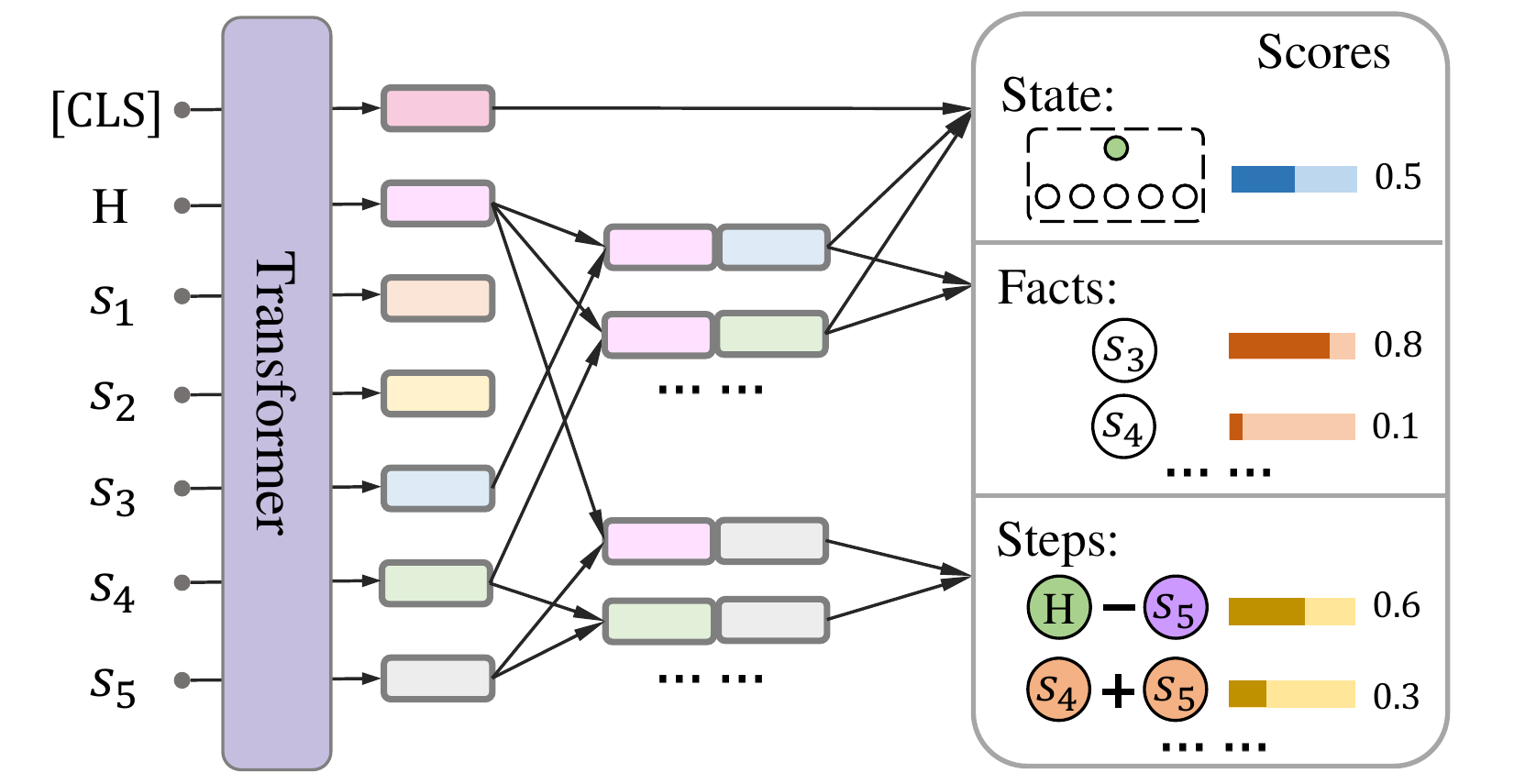}
\caption{Reasoning controller illustration.
Given a state, the controller predicts a score for the whole state, scores for facts, and scores for all possible steps.
}
\label{fig:controller}
\vspace{-5mm}
\end{figure}
}

\noindent\textbf{Encoding}.
We first encode the target hypothesis and facts of state with a pre-trained transformer: 
$\mathtt{[CLS]} H \mathtt{[SEP]} {s_1} \mathtt{[SEP]} \dots \mathtt{[SEP]} {s_n} \mathtt{[SEP]}.$
We obtain the contextualized representation $\bm{h}$ for $H$ and $\bm{f_i}$ for $s_i$ using the average contextualized representation of all tokens  within the sentence.

\noindent\textbf{Steps}.
We introduce feed forward networks $\operatorname{FFN_{ded}}$ and $\operatorname{FFN_{abd}}$ for deductive steps and abductive steps, respectively.
Each combination of two facts is a possible deductive step $(s_i, s_j)$.
Each combination of the target hypothesis and a fact is a possible abductive step $(H, s_k)$.
We score them by a score function $G_{step}$, 
{
\begin{equation}
\small
\begin{aligned}
\label{equ:step}
    G_{step}({s_i},{s_j}) & = \operatorname{FFN_{ded}}([\bm{f_i},\bm{f_j}]), \\ 
    G_{step}(\textit{H},{s_k}) & = \operatorname{FFN_{abd}}([\bm{h},\bm{f_k}]),
\end{aligned}
\end{equation}
}where $[\cdot]$ is the concatenate operation.
We normalize the step scores by applying $\operatorname{Softmax}$ over all possible deductive and abductive steps.

\noindent\textbf{Facts}.
The fact score indicates whether the fact is useful by how similar the fact is to the state's target hypothesis.
We assume that if a fact has a smaller depth in the gold entailment tree (i.e., closer to the root), it would be more similar to the target hypothesis than those facts with a larger depth.
We introduce $\operatorname{FFN_{fact}}$ as a learnable similarity function and determine the fact score by comparing it with the target,
{
\setlength{\abovedisplayskip}{6pt}
\setlength{\belowdisplayskip}{6pt}
\setlength{\abovedisplayshortskip}{6pt}
\setlength{\belowdisplayshortskip}{6pt}
\begin{equation}
    \small
    G_{fact}(s_i) = \sigma(\operatorname{FFN_{fact}}([\bm{h},\bm{f_i}])),
\end{equation}
}where $\sigma$ is the $\operatorname{Sigmoid}$ function.

\noindent\textbf{State}.
The state score reflects the quality of the current state and indicates whether this state should be used for further reasoning.
We assign the state score using the following two parts:
{
\setlength{\abovedisplayskip}{3pt}
\setlength{\belowdisplayskip}{3pt}
\setlength{\abovedisplayshortskip}{3pt}
\setlength{\belowdisplayshortskip}{3pt}
\begin{equation}
\label{equ:state}
    \small
    G_{state}(R) = \frac{\lambda}{n}\sum_{s_i \in S}
    G_{fact}(s_i) + (1-\lambda)\sigma(\operatorname{FFN_{cls}}(\bm{f_{\mathtt{[CLS]}}})),
\end{equation}
}where $\lambda$ is a learnable weight, $\bm{f_{\mathtt{[CLS]}}}$ is the representation of $\mathtt{[CLS]}$, $\operatorname{FFN_{cls}}$ is a feed forward network.
The first part helps choose states that contain more relevant facts and fewer distractors.
The second part comprehensively considers the whole state and gives the promising one a higher score.

\subsubsection{Controller Training}
\textbf{Training State Construction}.
\label{sec:state_construction}
We decompose the gold entailment trees into several intermediate states for training.
We add disturbances to the trees to make positive and negative states.
For each gold deductive step (e.g., $s_1\with s_2 \rightarrow i_1$), we use the deductive module to predict a conclusion $\hat{i}_1$.
If the predicted $\hat{i}_1$ is correct, we replace $i_1$ in the state with $\hat{i}_1$ to make new positive states.
Otherwise, we replace $i_1$ with $\hat{i}_1$ to make negative states.
The abductive modules are also used in a similar way.

\noindent \textbf{Loss Function}.
We train the controller with corresponding margin ranking losses $\mathcal{L}_{step},\mathcal{L}_{fact},$ and $\mathcal{L}_{state}$ to learn to rank the correct steps, facts, and states ahead of incorrect ones, respectively.
Specifically, the loss for scoring steps is 
{
\setlength{\abovedisplayskip}{3pt}
\setlength{\belowdisplayskip}{3pt}
\setlength{\abovedisplayshortskip}{3pt}
\setlength{\belowdisplayshortskip}{3pt}
\begin{equation}
    \small
    \mathcal{L}_{step} = \frac{1}{N_1} \sum_{(p^{+}, p^{-})} \varphi(G_{step}(p^{+}),G_{step}(p^{-}),m_{step}),
\end{equation}
}where $p^{+}$ and $p^{-}$ are the positive and negative step, $N_1$ is the number of ($p^{+}$, \!$p^{-}$) pairs, $\varphi(x_1,x_2,m) = \max(0, x_2-x_1+m)$ is the margin loss, and $m_{step}$ is the margin for steps.

\vspace{+1mm}
For facts, we have 
{
\setlength{\abovedisplayskip}{3pt}
\setlength{\belowdisplayskip}{3pt}
\setlength{\abovedisplayshortskip}{3pt}
\setlength{\belowdisplayshortskip}{3pt}
\small
\begin{align}
    \mathcal{L}_{fact} = & \frac{1}{N_2} \sum_{s^{+}_1, s^{+}_2\in S_{gold}} \varphi(G_{fact}(s^{+}_1),G_{fact}(s^{+}_2),m_{fact}) \nonumber \\  & + \frac{1}{N_3} \sum_{s^{-}\notin S_{gold}} -\log(1-G_{fact}(s^{-})),
\end{align}
}
$\!$where $s^{+}_1$ is a fact which has smaller depth in the gold tree than $s^{+}_2$, $s^{-}$ is the distractor, $N_2$ is the number of ($s^{+}_1$,$s^{+}_2$) pairs, $N_3$ is the number of distractors, and $m_{fact}$ is the margin for facts.

For states, we sample a positive state $R^{+}$ and a negative state $R^{-}$  from a tree and train the controller with
\begin{equation}
    \small
    \mathcal{L}_{state} = \varphi(G_{state}(R^{+}),G_{state}(R^{-}),m_{state}),
\end{equation}
where $m_{state}$ is the margin for states.

Finally, we average the above losses over all trees in the training split and train the controller with
\begin{equation}
    \small
    \mathcal{L} = \mathcal{L}_{step}+\mathcal{L}_{fact}+\mathcal{L}_{state}.
\end{equation}
Appendix~\ref{sec:controller_training} gives more controller training details.

\subsection{Reasoning Algorithm}
\label{sec:model_alg}
Since the entailment trees are generated iteratively and the search space for reasoning could be large for each iteration, we adopt beam search for efficient reasoning.
Given the initial state $H \Leftarrow S$, we first remove $s_i$ with a low \emph{fact score} to filter distractors.
Subsequently, we perform several reasoning iterations until the target hypothesis is proved or the maximum reasoning depth is reached.
In each iteration, we select the steps with the highest \emph{step scores}, execute the steps with all types of deductive or abductive modules, and construct novel states with the generated intermediate facts.
We remain the top-$K$ states ranked with \emph{state scores} for the next iteration, where $K$ is the beam size.
More algorithm details are in Appendix Algorithm~\ref{alg:reasoning}.

\section{Experiments}
We conduct experiments on EntailmentBank~\cite{DBLP:conf/emnlp/DalviJTXSPC21}, 
the first dataset supporting QA explanations in the form of the entailment tree.
EntailmentBank contains 1,840 entailment trees, each of which corresponds to a question from the ARC dataset~\cite{DBLP:journals/corr/abs-1803-05457}.
On average, each tree contains 7.6 nodes across 3.2 steps.
Summary statistics are shown in Table~\ref{tab:dataset}.

{\setlength{\abovecaptionskip}{2mm}
\begin{table}[t!]
\centering
\small
\begin{tabular}{@{}c|ccc|c@{}}
\toprule
 & Train & Dev & Test & All \\ \midrule
Questions / Trees & 1,131 & 187 & 340 & 1,840 \\
Entailment steps & 4,175 & 597 & 1,109 & 5,881 \\ \bottomrule
\end{tabular}
\caption{EntailmentBank statistics.}
\label{tab:dataset}
\vspace{-5mm}
\end{table}
}

\subsection{Evaluation Metrics}
Following~\citet{DBLP:conf/emnlp/DalviJTXSPC21}, we first align nodes in the predicted tree $T_{pred}$ with nodes in the gold tree $T_{gold}$ and then evaluate with three dimensions: 

\noindent $\bullet$ \textbf{Leaves:} 
To evaluate whether $T_{pred}$ uses the correct leaf facts, we compute \textbf{F1} score by comparing the predicted leaf facts $S_{pred}$ to $S_{gold}$.

\noindent $\bullet$ \textbf{Steps:} 
To evaluate whether the individual steps are \emph{structurally} correct, we compare all steps in two trees and compute \textbf{F1}.
A predicted step is considered \emph{structurally} correct if its children's identifiers (e.g., $s_{1}$, $i_{2}$) perfectly match the gold ones.

\noindent $\bullet$ \textbf{Intermediates:}
To evaluate whether the intermediate conclusions are correct, we report the \textbf{F1} of comparing the aligned intermediate conclusions.
A predicted intermediate sentence $\hat{i}$ is considered correct if the \texttt{BLEURT-Large-512} score of the aligned intermediate pair $(\hat{i},i)$ is larger than $0.28$\footnote{The threshold was picked using 300 manually labeled pairs~\cite{DBLP:conf/emnlp/DalviJTXSPC21}.}.

The \textbf{AllCorrect} score is 1 if F1 is 1, 0 otherwise\footnote{\label{footnote:debug}We repair a bug in the official evaluation code, which makes the Intermediate AllCorrect = 1 if the precision = 1 (rather than if F1 = 1), which leads to an overestimation on the Intermediate AllCorrect.}.
Given the above scores, we comprehensively evaluate $T_{pred}$ with \textbf{Overall AllCorrect} whose value is 1 if and only if all the leaves, steps and intermediates are all correct.
This is a strict metric since any error in $T_{pred}$ will lead to a score of 0.

\subsection{Baselines}
We compare with the SOTA entailment tree generation method \textbf{\EntialmentWriter}~\cite{DBLP:conf/emnlp/DalviJTXSPC21}, which directly generates the linearized trees (e.g., $s_2 \with s_5 \rightarrow i_1: \text{eruptions block sunlight}; s_4 \with i_1 \rightarrow H$) given $H+QA+S$ with an end-to-end encoder-decoder framework.
We also follow the ``Iterative'' ProofWriter~\cite{DBLP:conf/acl/TafjordDC21}, which is one of the SOTA proof generation methods for logical reasoning, to extend \EntialmentWriter to \textbf{\EntailmentWriterIter}.
\EntailmentWriterIter iteratively generates a part of the linearized tree in one forward process (e.g., $s_2 \with s_5 \rightarrow i_1:$ eruptions block sunlight;) and concatenates all parts to make the final tree.
It completes the step selection and entailment reasoning in a seq2seq model and does not provide the reasoning types of steps.

{\setlength{\abovecaptionskip}{2mm}
\begin{table*}[t!]
\small
\centering
\resizebox{\textwidth}{!}{%
\begin{tabular}{@{}clcccccccc@{}}
\toprule
\multirow{2}{*}{Task} & \multicolumn{1}{c}{\multirow{2}{*}{Method}} & \multirow{2}{*}{$n_{para}$} & \multicolumn{2}{c}{Leaves} & \multicolumn{2}{c}{Steps} & \multicolumn{2}{c}{Intermediates} & Overall \\
 & \multicolumn{1}{c}{} &  & F1 & AllCorrect & F1 & AllCorrect & F1 & AllCorrect & AllCorrect \\ \midrule
\multirow{5}{*}{\begin{tabular}[c]{@{}c@{}}Task1\\ (no-distractor)\end{tabular}} & \EntialmentWriter (T5-11B)\dag & 11.00 & 99.0 & 89.4 & 51.5 & 38.2 & 71.2 & \textbf{52.9}\dag & 35.6 \\
 & \EntialmentWriter (T5-large) & 0.77 & 98.4 & 84.1 & 50.0 & 38.5 & 67.0 & 35.9 & 34.4 \\
 & \EntailmentWriterIter (T5-large) & 0.77 & 99.8 & 97.6 & 51.6 & 38.5 & 68.3 & 36.5 & 35.0 \\ \cmidrule(l){2-10} 
 & \Oursep (Ours) & 0.22+6$\times$0.77 & \textbf{100.0} & \textbf{100.0} & \textbf{57.9} & \textbf{42.1} & \textbf{71.3} & 39.2 & \textbf{37.0} \\
 & \Ourpre (Ours) & 0.22+0.77 & \textbf{100.0} & \textbf{100.0} & 57.7 & 41.9 & 70.8 & 39.2 & 36.5 \\ \midrule
\multirow{5}{*}{\begin{tabular}[c]{@{}c@{}}Task2\\ (distractor)\end{tabular}} & \EntialmentWriter (T5-11B)\dag & 11.00 & \textbf{89.1} & \textbf{48.8} & 41.4 & 27.7 & \textbf{66.2} & \textbf{53.2}\dag & 25.6 \\
 & \EntialmentWriter (T5-large) & 0.77 & 83.2 & 35.0 & 39.5 & 24.7 & 62.2 & 28.2 & 23.2 \\
 & \EntailmentWriterIter (T5-large) & 0.77 & 85.2 & 40.9 & 38.9 & 26.8 & 63.5 & 29.1 & 25.0 \\ \cmidrule(l){2-10} 
 & \Oursep (Ours) & 0.22+6$\times$0.77 & 83.7 & 48.6 & \textbf{41.7} & \textbf{30.4} & 62.7 & 32.7 & \textbf{28.0} \\
 & \Ourpre (Ours) & 0.22+0.77 & 82.7 & 46.1 & 41.3 & 29.6 & 61.4 & 32.4 & 27.7 \\ \midrule
\multirow{5}{*}{\begin{tabular}[c]{@{}c@{}}Task3\\ (full-corpus)\end{tabular}} & \EntialmentWriter (T5-11B)\dag & 11.00 & \textbf{39.7} & 3.8 & 7.8 & 2.9 & 36.4 & 13.2\dag & 2.9 \\
 & \EntialmentWriter (T5-large) & 0.77 & 30.9 & 1.2 & 4.4 & 1.2 & 28.8 & 5.6 & 1.2 \\
 & \EntailmentWriterIter (T5-large) & 0.77 & 32.4 & 1.8 & 4.4 & 1.5 & 29.7 & 6.5 & 1.5 \\ \cmidrule(l){2-10} 
 & \Oursep (Ours) & 0.22+6$\times$0.77 & {34.8} & \textbf{8.7} & \textbf{9.8} & \textbf{8.6} & \textbf{36.7} & \textbf{20.4} & \textbf{8.6} \\
 & \Ourpre (Ours) & 0.22+0.77 & {34.8} & \textbf{8.7} & \textbf{9.8} & \textbf{8.6} & 36.6 & \textbf{20.4} & \textbf{8.6} \\ \bottomrule
\end{tabular}
}
\caption{
Automatic evaluation results on the EntailmentBank test split.
\dag\xspace indicates results from the published paper\textsuperscript{\ref{footnote:debug}}.
$n_{para}$ denotes the number of model parameters (B).
}
\label{tab:main_result}
\vspace{-3mm}
\end{table*}
}

{\setlength{\abovecaptionskip}{2mm}
\begin{table}[t!]
\small
\centering
\resizebox{\columnwidth}{!}{%
\begin{tabular}{@{}lcccc@{}}
\toprule
  & \multicolumn{2}{c}{Task1} & \multicolumn{2}{c}{Task2} \\
 Method & Automatic & Manual & Automatic & Manual \\ \midrule
\EntialmentWriter (T5-large) & 35 & 46 & 21 & 26 \\
\EntailmentWriterIter (T5-large) & 35 & 47 & 25 & 35 \\ \midrule
\Ourpre (Ours) & \textbf{36} & \textbf{53} & \textbf{27} & \textbf{39} \\ \bottomrule
\end{tabular}
}
\caption{Entailment tree evaluation results on 100 uniformly sampled questions from the test split.
We report the proportion (\%) of the predicted trees that are rated as valid, following automatic and manual evaluation.}
\label{tab:tree_manual}
\vspace{-2mm}
\end{table}
}

\subsection{Implementation Details}
\textbf{Modules.}
We implement the entailment modules on top of T5-large~\cite{DBLP:journals/jmlr/RaffelSRLNMZLL20} with the following two implementations.
(1) \textbf{Separated.} 
We implement each module separately.
We have six models in total, corresponding to the three reasoning types of deductive and abductive versions.
(2) \textbf{Prefixed.} 
We implement all modules with a single model.
To specify which reasoning type the model should perform, we follow \citet{DBLP:journals/jmlr/RaffelSRLNMZLL20} to add a type-specific prefix (e.g., ``deductive substitution:'') to the input before feeding it to the model.
To evaluate the modules, we annotate the types of 275 steps in the dev split.
We train the modules with a batch size of 20 for 100 epochs.

\noindent \textbf{Controller.}
The controller is implemented with albert-xxlarge-v2~\cite{DBLP:journals/corr/abs-1909-11942}.
We train two individual controllers for Task1 and Task2.
For Task3, we reuse the Task2 model without additional training.
The controllers are trained with a batch size of 10 for 1,000 epochs.
The margins $m_{step}$, $m_{fact}$, and $m_{state}$ are tuned on the development split and all set to 0.1.

\noindent\textbf{Algorithm.}
For Task1, we iterate until all facts in $S$ are used.
For Task2, we use a fact score threshold of 0.001 to filter distractors and a maximum reasoning depth of 5.
We select the top 10\% steps for each state and set the beam size to 10.
All hyper-parameters are selected using the dev split (Appendix~\ref{sec:alg_parameter}).
For Task3, we follow \citet{DBLP:conf/emnlp/DalviJTXSPC21} to retrieve 25 sentences from the corpus $C$ using the $H$ as the query.
We use the same retrieval results as EntailmentWriter for a fair comparison. 
Model checkpoints are selected using the dev split.
More implementation details can be found in the Appendix.

\section{Result Analysis}
\subsection{Automatic Evaluation}
As shown in Table~\ref{tab:main_result}, our methods outperform all baseline methods on the strictest metric Overall AllCorrect for all three tasks.
Notice that the trees generated by our methods only contain 2-premise steps, which would lead to a 0 Overall AllCorrect score on 26\% of test samples whose annotations contain $n$-premise ($n > 2$) steps.
Even so, our \Oursep still obtains an absolute improvement of 1.4\%/2.4\%/5.7\% on Task1/2/3 in comparison to the strongest baseline.
With only 9.0\% of the model parameters, \Ourpre can outperform the \EntialmentWriter (T5-11B) by absolute 0.9\%/2.1\%/5.7\% on Task1/2/3.
In the case of using a comparable amount of model parameters, \Ourpre also outperforms the \EntailmentWriterIter (T5-large) by a large margin.
For Task3, we note that all methods perform poorly.
The main reason is that the retrieved facts may not contain all the required facts $S_{gold}$ (68\% of the cases).
We note that \modelname underperforms the baselines on some metrics, probably due to the inaccuracy of the tree alignment algorithm in the automatic evaluation (Appendix~\ref{Sec:discussion_auto_eval}).

{\setlength{\abovecaptionskip}{1.5mm}
\begin{figure}[t!]
\centering
\includegraphics[width=\columnwidth]{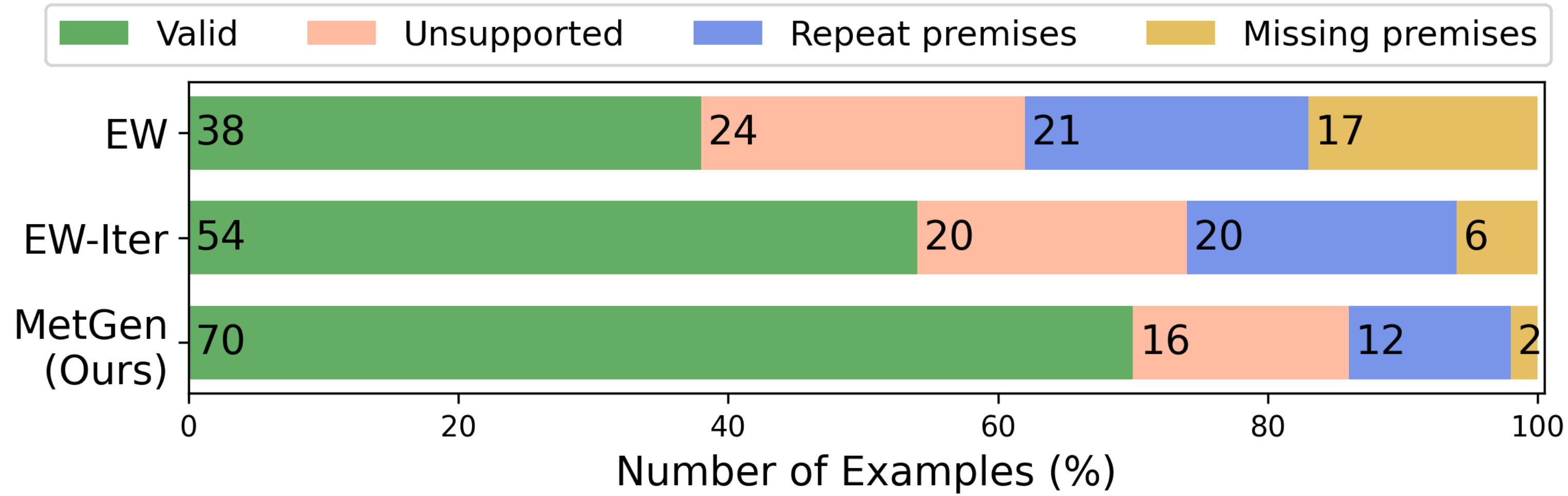}
\caption{Manual evaluation results of 100 single-step entailments uniformly sampled from the predicted trees of Task2 test spilt.
EW denotes EntailmentWriter.
}
\label{fig:step_manual}
\vspace{-5mm}
\end{figure}
}

\subsection{Manual Evaluation}
\label{sec:manual_eval}
As analysed by~\citet{DBLP:conf/emnlp/DalviJTXSPC21}, the automated metrics might misjudge some valid trees and thus underestimate the performance.
To make a more accurate comparison, we perform the manual evaluation.
We compare three methods with a comparable amount of model parameters, \EntialmentWriter (T5-large), \EntailmentWriterIter (T5-large), and \Ourpre.
For each step and tree, we invite three students as experts to evaluate the validity.
The inter-annotator agreement (Cohen's kappa statistic) is 0.85/0.76 for the step/tree, indicating the substantial agreement between annotators.

\noindent\textbf{Validity of Full Entailment Trees.} 
As shown in Table~\ref{tab:tree_manual}, under the manual evaluation, \modelname outperforms the baselines with large margins.

\noindent\textbf{Validity of Individual Entailment Steps.}
We review the validity of the single-step entailment and annotate each step with one of the four categories:

\noindent {$\bullet$} \textbf{Valid:} 
The step conclusion can be inferred from the premises and does not trivially repeat them.

\noindent {$\bullet$} \textbf{Unsupported:} 
The conclusion is in conflict with, irrelevant with, or not followed from the premises.

\noindent {$\bullet$} \textbf{Repeat premises:} 
The conclusion trivially repeats one or more of the premises.

\noindent {$\bullet$} \textbf{Missing premises:}
The conclusion uses knowledge unstated in the premises.
The step would be correct if one additional premise from $S$ is added.

As shown in Figure~\ref{fig:step_manual}, \modelname achieves considerable improvement in the validity of steps compared to the baseline methods.
We note that 17\% of the steps of \EntialmentWriter belong to missing premises.
\modelname constrains the reasoning types of steps and uses the premise-related and context-independent entailment modules to perform every single step.
This can reduce the cases of missing premises (from 17\% to 2\%) and improve the validity of the conclusions (from 38\% to 70\%).

{\setlength{\abovecaptionskip}{2mm}
\begin{table}[t!]
\small
\centering
\setlength\tabcolsep{1pt}
\resizebox{\columnwidth}{!}{%
\begin{tabular}{ccccc|cc}
\toprule
\multirow{2}{*}{} &
  \multirow{2}{*}{\begin{tabular}[c]{@{}c@{}}Implem- \\ entation\end{tabular}} &
  \multirow{2}{*}{Models} &
  \multirow{2}{*}{\begin{tabular}[c]{@{}c@{}}Reasoning \\ Type\end{tabular}} &
  \multirow{2}{*}{\begin{tabular}[c]{@{}c@{}}Training\\ Data\end{tabular}} &
  \multirow{2}{*}{\begin{tabular}[c]{@{}c@{}}Overall\\ AllCorrect\end{tabular}} &
  \multirow{2}{*}{\begin{tabular}[c]{@{}c@{}}Single-step\\ Accuracy\end{tabular}} \\
    &     &            &   &     &      &      \\ \midrule
(a) & Sep & 6$\times$T5-large   & \checkmark & S+E & \textbf{28.0} & \textbf{81.0} \\
(b) & Sep & 6$\times$BART-large & \checkmark & S+E & 26.2 & 77.0 \\
(c) & Sep & 6$\times$T5-base    & \checkmark & S+E & 27.3 & 78.0 \\
(d) & Sep & 6$\times$T5-large   & \checkmark & E   & 27.8 & 79.5 \\
(e) & Sep & 6$\times$T5-large   & \checkmark & S   & 23.5 & 43.6 \\
(f) & Pre & 1$\times$T5-large   & \checkmark & S+E & 27.7 & 78.4 \\
(g) & Pre & 1$\times$T5-large   & \checkmark & E   & 27.4 & 78.1 \\
(h) & Pre & 1$\times$T5-large   & $\times$   & E   & 25.9 & 76.0 \\ \bottomrule
\end{tabular}%
}
\caption{
Ablation results on entailment modules.
Sep/Pre indicates seperated/prefixed.
S/E denotes the synthesis/EntailmentBank step training data.
}
\label{tab:ablation_module}
\end{table}
}

\subsection{Ablation Study}
\textbf{Entailment Modules Analysis.}
Table~\ref{tab:ablation_module} reports the ablation results on modules.
We report the Overall AllCorrect on test spilt and the single-step entailment accuracy 
on the labeled dev steps, and can make the following observations.
(1) \textbf{Separated\!\! vs.\! Prefixed.\!}
We can see that \Ourpre achieves slightly worse performance than \Oursep ((a) vs. (f) and (d) vs. (g)).
This is mainly because separate modules could better learn different types of reasoning.
However, in our final system, we still choose to use \Ourpre due to the consideration of model size.
(2) \textbf{Clarifying Reasoning Types.}
We train a module to infer without distinguishing or assigning specific reasoning types.
We find that the performance drops from 27.4\% to 25.9\% ((g) vs. (h)), suggesting that clarifying the reasoning types of the entailment steps is crucial for generating entailment trees.
(3) \textbf{Training Data.}
Comparing (a) and (d), we find that training with the synthesis data could improve the accuracy.
Without tuning on EntailmentBank (setting (e)), the modules might not adapt to the science domain and obtain low step accuracy.
However, the well-trained controller would verify and filter the error conclusions, thus our method can still achieve 23.5\% on Overall AllCorrect.
(4) \textbf{Generative Model.}
A stronger generative model, which achieves higher single-step accuracy, could achieve higher tree generation performance (comparing (a), (b) and (c)), indicating that our method can be further improved with stronger entailment modules.

\noindent \textbf{Controller and Algorithm Analysis.}
(1) \textbf{Is the reasoning controller necessary?}
To answer this question, we design a \textbf{heuristic} generation algorithm without the controller (Appendix~\ref{sec:heuristic_search}).
It uses the \texttt{BLEURT} scores as heuristic information to guide the reasoning.
As shown in Table~\ref{tab:ablation_controller}, the heuristic method achieves observable lower performance.
The controller could aid in eliminating the error steps and states, so as to find the valid trees efficiently and accurately.
Without the controller, we find it difficult to find effective heuristic information.
(2) \textbf{Effect of Abductive Steps.}
The generation performance drops when abductive steps are not used.
This suggests that abductive steps, as a way of backward searching, could help improve the quality of generated trees.

{\setlength{\abovecaptionskip}{2mm}
\begin{table}[t]
\small
\centering
\resizebox{\columnwidth}{!}{%
\begin{tabular}{@{}clcccc@{}}
\toprule
Task & Method & Leaves & Steps & Intermediate & Overall \\ \midrule
\multirow{3}{*}{Task1} & controller & \textbf{100.0} & \textbf{42.1} & \textbf{39.2} & \textbf{37.0} \\
 & \quad w/o abduction & 100.0 & 41.4 & 38.4 & 36.2 \\
 & heuristic & 100.0 & 31.2 & 31.2 & 28.8 \\ \midrule
\multirow{3}{*}{Task2} & controller & \textbf{48.6} & \textbf{30.4} & \textbf{32.7} & \textbf{28.0} \\
 & \quad w/o abduction & 44.5 & 28.3  & 31.6 & 27.0 \\
 & heuristic & 3.2 & 3.2 & 12.1 & 3.2 \\ \bottomrule
\end{tabular}
}
\caption{Ablation results on the reasoning controller. We report the  AllCorrect scores on the test split.
}
\label{tab:ablation_controller}
\end{table}
}

{\setlength{\abovecaptionskip}{2mm}
\begin{figure}[t!]
\centering
\includegraphics[width=\columnwidth]{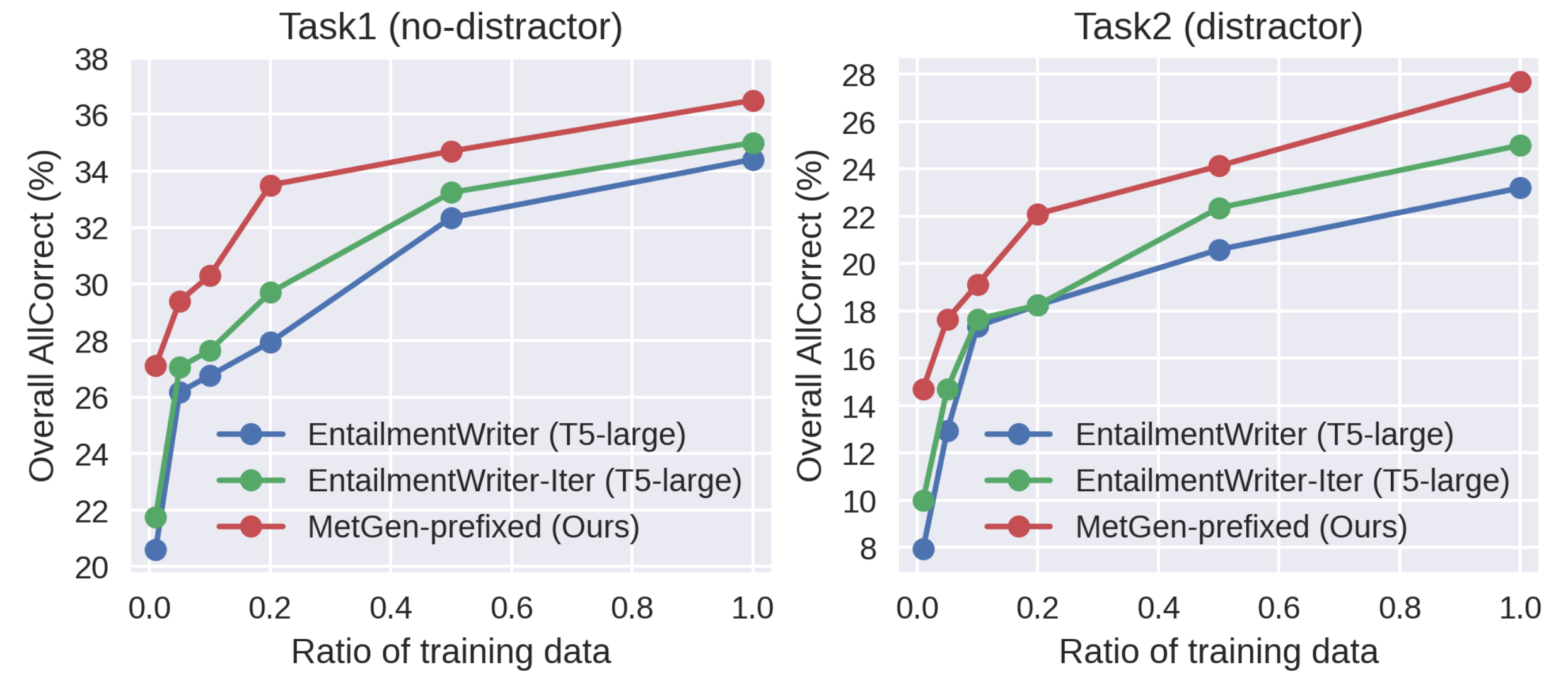}
\caption{Results on different ratios (0.01, 0.05, 0.10, 0.20, 0.50, 1.00) of EntailmentBank training data.
}
\label{fig:few-shot}
\end{figure}
}

{\setlength{\abovecaptionskip}{2mm}
\begin{table}[t!]
\small
\centering
\resizebox{\columnwidth}{!}{%
\begin{tabular}{@{}lcccc@{}}
\toprule
 & \multicolumn{2}{c}{eQASC} & \multicolumn{2}{c}{eOBQA} \\
 Method & P@1 & NDCG & P@1 & NDCG \\ \midrule
\EntialmentWriter (T5-large) & 52.48 & 73.14 & 69.07 & 89.05 \\
\EntailmentWriterIter (T5-large) & 52.56 & 73.28 & 72.15 & 90.19 \\ \midrule
\Ourpre (Ours) & \textbf{55.81} & \textbf{74.19} & \textbf{74.89} & \textbf{90.50} \\ \bottomrule
\end{tabular}
}
\caption{\rw{Cross-dataset} results on the eQASC and eOBQA test split.}
\label{tab:eqasc}
\end{table}
}

\subsection{Data-scarce Setting}
Figure~\ref{fig:few-shot} reports the results in the data-scarce setting.
Our method is more data-efficient.
With only 1\% of the EntailmentBank training data, our method obtains 14.7\% on Task2 Overall AllCorrect, in comparison to 10.0\% of the strongest baseline.
When the data is scarce, the advantage of training our modules with synthetic data becomes more significant.
It can help alleviate the overfitting on few EntailmentBank sentences.

\subsection{\rw{Cross-dataset Setting}}
To test the generalization capability of our method, we conduct \rw{cross-dataset} experiments on datasets eQASC and eOBQA~\cite{DBLP:conf/emnlp/JhamtaniC20}, which collect \emph{one-step} entailment trees for questions from QASC~\cite{DBLP:conf/aaai/KhotCGJS20} and OpenBookQA~\cite{DBLP:conf/emnlp/MihaylovCKS18}, respectively.
Given $H$ and $S$, their task requires selecting the valid one-step trees (e.g., $s_1 \with s_2 \rightarrow H$) from a candidate set.
We apply the Task2 models (without fine-tuning on eQASC or eOBQA) to select from the candidate trees (Appendix~\ref{sec:eqasc}).
Following \citet{DBLP:conf/emnlp/JhamtaniC20}, we evaluate models with the \textbf{P@1} and \textbf{NDCG} metrics.
Questions with no valid tree are filtered.
As shown in Table~\ref{tab:eqasc}, our method achieves better generalization performance.
Instead of training a seq2seq model with a single generation loss, our method explicitly models the step and state selection ability (equation~(\ref{equ:step}) and~(\ref{equ:state})) and guides the controller with specific losses to rank the correct ones ahead of incorrect ones.
Such a manner could aid in alleviating the overfitting on training data and improve the generality.

\section{Conclusion}
We propose \modelname, a module-based framework to generate the entailment trees for explaining answers.
\modelname reasons with single-step entailment modules and the reasoning controller.
Experiments on EntailmentBank benchmark show \modelname can generate valid trees with reliable steps and achieve SOTA performance.

\section{Acknowledgements}
We appreciate the anonymous reviewers for their insightful comments.
We would like to thank Zhihong Shao for the helpful discussions.
This work is supported by the National Key Research and Development Program of China (No. 2018AAA0100701), and the Guoqiang Institute of Tsinghua University with Grant No. 2020GQG0005.


\bibliography{anthology,custom}
\bibliographystyle{acl_natbib}

\newpage
\clearpage

\appendix

\section{Entailment Modules Training Details}
\subsection{Synthetic Data} 
\label{sec:parapattern_data}

We follow the ParaPattern~\cite{DBLP:conf/emnlp/BostromZCD21} to collect synthetic training data for the entailment modules.
Since they only consider the substitution and contraposition deductions, we extend the method to conjunction and if-then deductions by designing the specific syntactic templates and construction rules.
Table~\ref{tab:module_pattern} shows the used syntactic patterns.
We use Spacy\footnote{https://spacy.io/} to match sentences from Wikipedia (version ``20200501.en'').
In total, we collect about 24k, 443k, and 97k sentences for substitution, conjunction, and if-then modules, respectively.
We follow~\citet{DBLP:conf/emnlp/BostromZCD21} to train the modules on the synthetic data with a learning rate of 3e-5 for 1 epoch.

\subsection{Reasoning Type Annotations of EntailmentBank}
The original steps in the EntailmentBank are not annotated with reasoning types.
We manually annotated the reasoning types of 400 steps in the training split (Train-manual) and 275 steps in the development split (Dev-manual).
To label the remaining steps in the training split, we train a classifier with the Train-manual steps.
We use the Roberta-large~\cite{DBLP:journals/corr/abs-1907-11692} as our classifier.
It achieves an accuracy rate of 88\% on the Dev-manual steps.
We use the classifier to predict the reasoning types of the remaining 2-premise steps and take the predicted types as the pseudo labels (Train-pseudo).
Table~\ref{tab:step_type_anno} shows the statistics of the reasoning type annotations.

\begin{table}[h]
\centering
\small
\begin{tabular}{@{}lcccc@{}}
\toprule
Split & Sub. & Conj. & If-then & All  \\ \midrule
Train-manual & 211  & 105   & 84      & 400  \\
Train-pseudo & 2,441 & 812   & 535     & 3,788 \\
Dev-manual   & 153  & 71    & 51      & 275  \\ \bottomrule
\end{tabular}
\caption{Statistics of the step reasoning type annotations.}
\label{tab:step_type_anno}
\end{table}

\section{Controller Training Details}
\label{sec:controller_training}

\noindent \textbf{Training Data.}
We decompose the gold entailment trees into several intermediate states for training.
For example, the tree in Figure~\ref{fig:task_intro}(c) can be decomposed into the following positive states: 
$R_0: H\Leftarrow \{s_1,s_2,s_3,s_4,s_5\}$, $R_1: H\Leftarrow \{s_1,s_3,s_4,i_1\}$, and $R_2: i_1\Leftarrow \{s_1,s_2,s_3,s_5\}$.
The state $R_0$ has two distractors $s_1$ and $s_3$, one positive deductive step $s_2\with s_5 \rightarrow i_1$, and one positive abductive step $H - s_4 \rightarrow i_1$.
We add disturbances to the trees to make positive and negative states.
For the state $R_1$, the fact $i_1$ is the conclusion of gold step $s_2\with s_5 \rightarrow i_1$.
We use a deductive module to predict a conclusion $\hat{i}_1$ given $s_2$ and $s_5$.
If the predicted $\hat{i}_1$ is correct, we replace $i_1$ with $\hat{i}_1$ to make new positive states $R_1^{+}: H\Leftarrow \{s_1,s_3,s_4,\hat{i}_1\}$.
The $R_1^{+}$ can be used to perform further reasoning.
Otherwise, we replace $i_1$ with $\hat{i}_1$ to make negative states $R_1^{-}$.
The $R_1^{-}$ contains an incorrect conclusion $\hat{i}_1$ and thus should not be used for further reasoning.
The reasoning controller should be trained to learn to distinguish between $R_1^{+}$ and $R_1^{-}$ and give the $R_1^{+}$ a higher state score than $R_1^{-}$.
To judge whether the generated $\hat{i}_1$ is correct, we follow the evaluation metrics~\cite{DBLP:conf/emnlp/DalviJTXSPC21} to use \texttt{BLEURT}.
The predicted $\hat{i}_1$ is considered correct if the \texttt{BLEURT} score between $\hat{i}_1$ and the gold $i_1$ is larger than 0.28.

\section{Reasoning Algorithm and Hyperparameter Analysis}
\label{sec:alg_parameter}
\begin{algorithm}[h!]
\caption{Reasoning Algorithm}
\label{alg:reasoning}
\textbf{Input:} Hypothesis $H$, fact sentences $S$, controller, deductive modules $M_{ded}$, abductive modules $M_{abd}$ \\
\textbf{Parameter:} Beam size $K$, max reasoning depth $D$, distractor threshold $\theta$, step sampling rate $\tau$
\begin{algorithmic}[1] 
\STATE \textrm{// Construct initial reasoning state}
\STATE Remove $s_i$ with \emph{fact score} less than $\theta$ in $S$
\STATE $R_{init} \gets$ ($H \Leftarrow$ \textrm{the filtered sentences} $S'$)
\STATE $\mathcal{R}_{beam} \gets \{ R_{init} \}, \mathcal{R} \gets \mathcal{R}_{beam}$

\STATE \textrm{// Reasoning with beam search}
\WHILE{\textrm{the depth does not reach $D$}}
\STATE $\mathcal{R}'_{beam} \gets \{\}$
    \FOR{$R \in \mathcal{R}_{beam}$}
        \STATE \textrm{// Select promising steps}
        \FOR{$p \in$ steps of $R$ with top $\tau\%$ \emph{step score}}
            \STATE \textrm{// Single-step entailment reasoning}
            \FOR{$m \in M_{ded}$ or $m \in M_{abd}$}
                \STATE execute step $p$ with module $m$ and obtain a novel intermediate fact $i$
                \STATE construct a new state $R_{new}$ with the step $p$ and the fact $i$
                \STATE $R'_{beam} \gets R'_{beam} \cup \{R_{new}\}$
            \ENDFOR
        \ENDFOR
    \ENDFOR
    \STATE \textrm{// Verify and select states}
    \STATE $\mathcal{R}_{beam} \gets$ $K$ states with the highest \emph{state scores} from $R'_{beam}$
    \STATE $\mathcal{R} \gets\mathcal{R} \cup \mathcal{R}_{beam}$
\ENDWHILE

\STATE \textrm{// Construct the entailment tree}
\FOR{$R \in \mathcal{R}$}
    \STATE Align the target of $R$ to the most similar fact sentence of $R$ to make a tree $T$
\ENDFOR
\STATE Select the tree $\hat{T}$ with highest score
\STATE \textbf{Return} The entailment tree $\hat{T}$ 
\end{algorithmic}
\end{algorithm}

Algorithm~\ref{alg:reasoning} shows the whole reasoning process.
The hyperparameters are selected with the development split, as shown in Figure~\ref{fig:parameter_search}.
We select a beam size of 10, a max reasoning depth of 5, a distractor threshold of 0.001, and a step sampling rate of 10\%.
We only consider the steps whose sentences have word overlap.
When constructing the entailment tree, 
we use the \texttt{BLEURT} scores to align the target of a state to the most similar fact.
Note that when making a new reasoning state with the step $p$ and the novel intermediate fact $i$, if the step $p$ is a backward abductive step, we replace the original target hypothesis with $i$ and treat the $i$ as the target hypothesis which the new state aims to prove (as shown in Figure~\ref{fig:search}).
We run our method three times and report the average performance.

\section{Heuristic Reasoning Algorithm without the Controller}
\label{sec:heuristic_search}
To investigate the effect of the reasoning controller for entailment tree generation, we design a heuristic generation algorithm that does not use the reasoning controller.
Since the cost of traversing the entire search space is unaffordable, we adopt the beam search.
In each reasoning state, we try all possible steps with entailment modules and make new candidate reasoning states.
To select the correct states, we use the \texttt{BLEURT} scores as the heuristic information to guide the search process.
Specifically, given a candidate state $R: H \Leftarrow S$, we estimate the similarity between a fact $s_i \in S$ and the target $H$ by
\begin{equation}
    \small
    G'_{fact}(s_i) = \texttt{BLEURT}(H,s_i),
\end{equation}
and then score a candidate state by
\begin{equation}
    \small
    G'_{state}(R) = \frac{1}{n} \sum _{s_i \in S} G'_{fact}(s_i).
\end{equation}
The top-$K$ candidate states with the highest state scores are selected to perform further reasoning, where $K$ is the beam size.
We use the same beam size as the algorithm with the controller uses.

{\setlength{\abovecaptionskip}{2mm}
\begin{table*}[t!]
\centering
\includegraphics[width=0.95\textwidth]{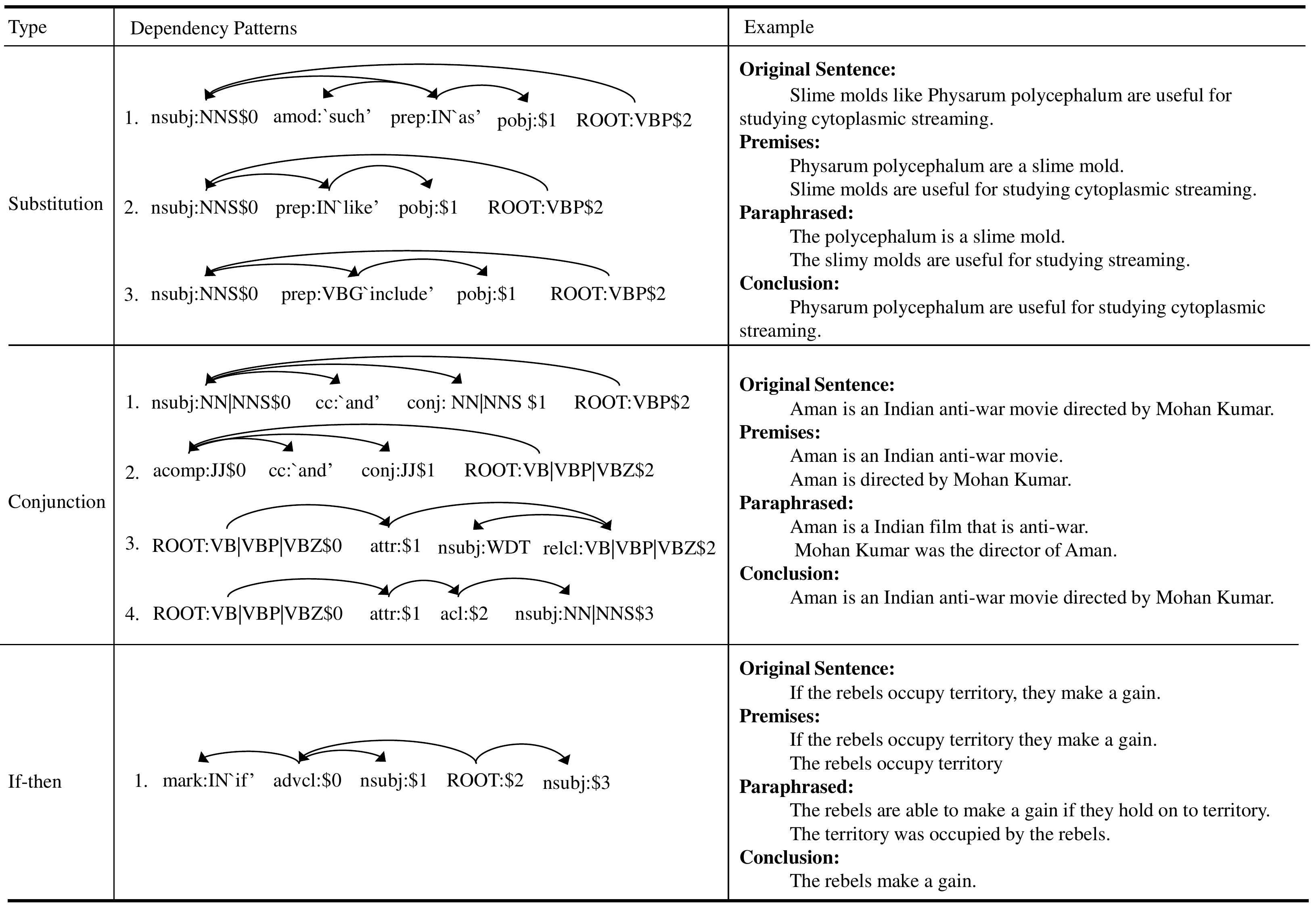}
\caption{
The syntactic patterns used on data scraping and the training examples for deductive entailment modules.
Pattern nodes are donated as $\operatorname{dep:POS`lemma'}\$i$, where $\operatorname{dep}$ contains the dependency relations of the matching token, $\operatorname{POS}$ contains the part-of-speech tags of the matching token, $\operatorname{`lemma'}$ contains the lemmatized form of the matching token, and $\$i$ indicates that a matching token and its subtree will be used as a match variable for rule-based rewriting. $|$ means ``or''.
}
\label{tab:module_pattern}
\vspace{-3mm}
\end{table*}
}

{\setlength{\abovecaptionskip}{2mm}
\begin{figure*}[t!]
\centering
\includegraphics[width=0.95\textwidth]{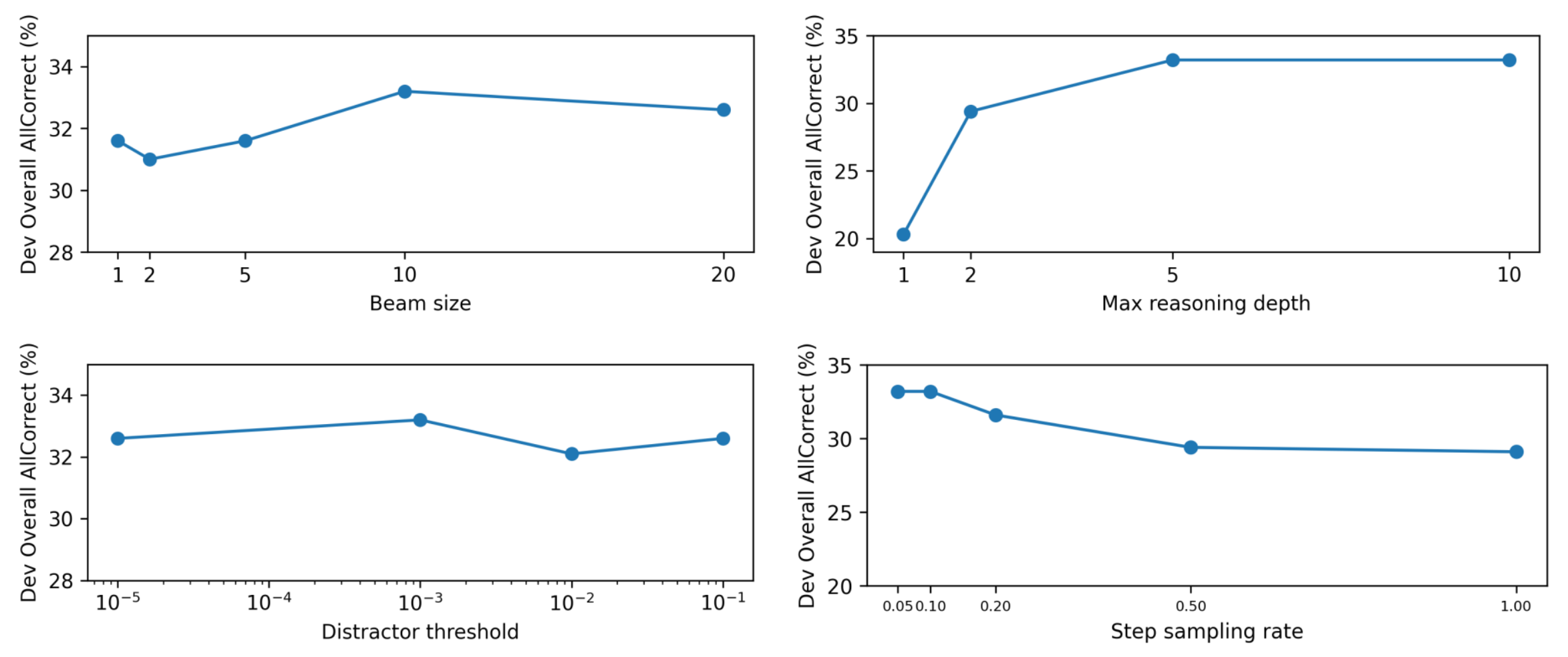}
\caption{Hyperparameter analysis on the Task2 development spilt.
}
\label{fig:parameter_search}
\vspace{-5mm}
\end{figure*}
}

\section{Experiment Details on eQASC and eOBQA}
\label{sec:eqasc}
For each question+answer pair, the eQASC/eOBQA provides the corresponding hypothesis $H$, about 10/4 facts as $S$, and a candidate set of steps.
Each candidate step is a 2-premise single step from two facts to $H$ (e.g., $s_1 \with s_2 \rightarrow H$) and can be viewed as a \emph{one-step entailment tree} with three nodes.
The target is to select the correct trees/steps from the candidate set.
There might be more than one correct tree in the candidate set.
We conduct experiments on the questions with at least one correct entailment tree (677 eQASC questions and 79 eOBQA questions).
Since the given $S$ contains distractors, we adopt the Task2 models trained on EntailmentBank (without further fine-tuning on eQASC and eOBQA) to perform cross-dataset experiments.

For our method, we follow our Task2 reasoning algorithm to select from the candidate trees/steps.
Specifically, we first filter out the facts in $S$ with low fact scores using a threshold (selected using the development split).
Then we predict the step scores for the candidate steps and select the step with the highest score.
For the EntailmentWriter, we feed the $S$ and $H$ to the EntailmentWriter and score each candidate step with the $\frac{1}{\operatorname{PPL}}$, where ${\operatorname{PPL}}$ is the perplexity of the sequence segment representing the step (e.g., $sent1\ \& \ sent2$ for $s_1 \with s_2$ in the official EntailmentWriter implementation).

We follow the official evaluation metrics of eQASC and eOBQA.
The P@1 (Precision@1) measures the fraction of cases where the selected tree (topmost ranked) is correct.
It is equivalent to the Overall AllCorrect score between the top-1 predicted one-step tree and the best-matching gold tree.
The NDCG (Normalized Discounted Cumulative Gain) metric measures how well ranked the candidate trees are when ordered by the predicted scores.
It reflects the model's ability to distinguish the validity of trees and rank the correct trees ahead of the incorrect ones.

\section{Main Experimental Environments}
We deploy all models on a server with 500GB of memory and one 40G A100 GPU.
Specifically, the configuration environment of the server is ubuntu 21.04 and our code mainly depends on python 3.8.10 and PyTorch 1.7.1.
We use the pre-trained language models from \texttt{HuggingFace Transformers}\footnote{https://github.com/huggingface/transformers}.
We use the Adafactor optimizer~\cite{DBLP:conf/icml/ShazeerS18} implemented by \texttt{HuggingFace Transformers}.

\begin{figure*}[t]
\centering
\includegraphics[width=\textwidth]{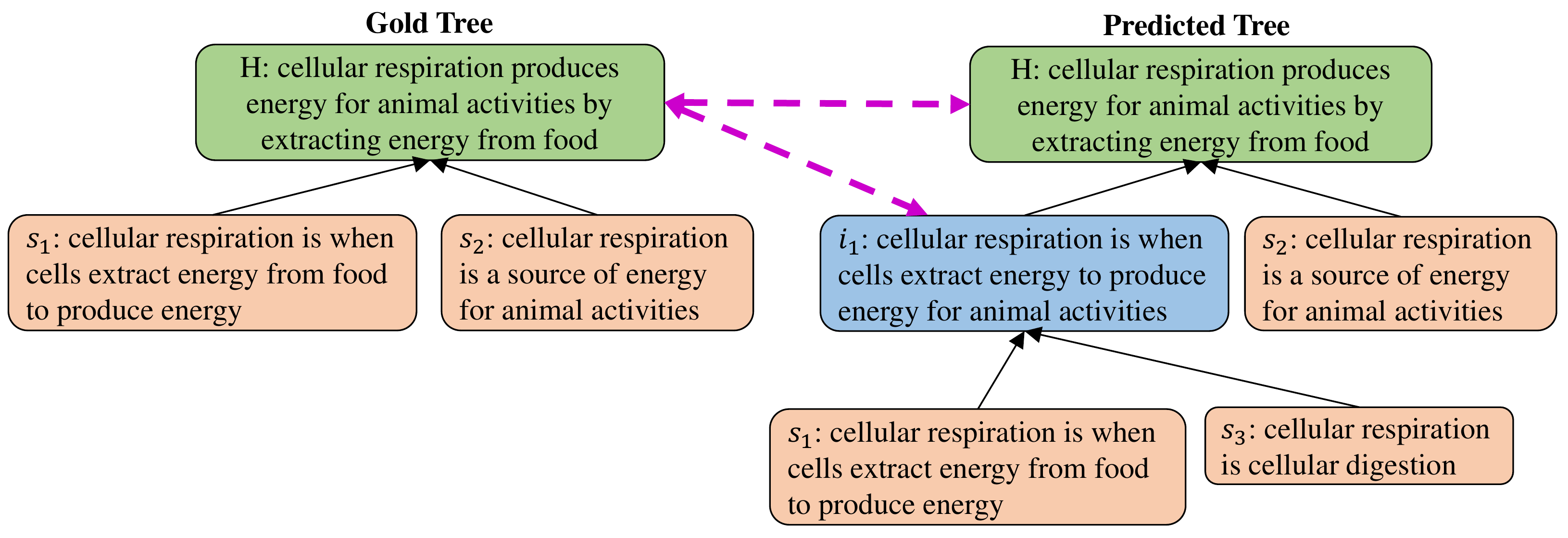}
\caption{An example case illustrating the potential inaccuracy of the automatic evaluation metrics.
In the predicted tree, the fact $s_3$ is a distractor and the step $s_1 \with s_3 \rightarrow i_1$ is not a valid entailment.
Following the official evaluation code, the nodes $i_1$, $H$ in the predicted tree are aligned to the $H$, $H$ in the gold tree, respectively (the dotted line).
By comparing the aligned intermediate nodes ($i_1$ vs. $H$, $H$ vs. $H$), the predicted tree achieves a Step F1 score of 0.0 and an Intermediate F1 score of 1.0.
The Intermediate F1 score being 1.0 should have indicated that the predicted tree has perfect intermediate conclusions.
However, the $i_1$ is not entailed by the $s_1$ and $s_3$.
}
\label{fig:metric_fail_case}
\end{figure*}

\section{Discussion on the Automatic Evaluation}
\label{Sec:discussion_auto_eval}
As discussed by~\citet{DBLP:conf/emnlp/DalviJTXSPC21}, the automatic entailment tree evaluation metrics might misjudge in some cases (e.g., tree structure variation) and still need to be improved.
In fact, how to quantitatively evaluate a predicted tree remains a challenging problem.
In the existing metric, the first step is the tree alignment algorithm~\cite{DBLP:conf/emnlp/DalviJTXSPC21}.
The nodes in the predicted tree $T_{pred}$ are aligned to the nodes in the gold tree $T_{gold}$ for further comparison.
Each non-leaf node $i_{pred}$ of $T_{pred}$ is aligned to the first non-leaf node $i_{gold}$ where the Jaccard similarity of their respective leaf sentences is maximum.
For any $i_{pred}$ with zero Jaccard similarity to all gold nodes, it is aligned to a dummy gold node with a blank conclusion.
In the official implementation, 
(1) each $i_{gold}$ may correspond to more than one $i_{pred}$, while there is no penalty for duplication when calculating Intermediate F1;
(2) the root node (the \emph{given} hypothesis sentence which is \emph{identical} in $T_{pred}$ and $T_{gold}$) is trivially viewed as a normal intermediate node (the \emph{novel generated} intermediate sentence).
Because of these two reasons, the Intermediate F1 might achieve a high score (indicating the $T_{pred}$ can draw correct intermediate conclusions from the premises), even when the Step F1/AllCorrect is relatively low (indicating the $T_{pred}$ does not select the correct premises for the intermediate nodes).
For example, the EntailmentWriter (T5-11B) for Task3 achieves an Intermediate F1 of 36.4\% while the Step F1/AllCorrect is only 7.8\%/2.9\%~\cite{DBLP:conf/emnlp/DalviJTXSPC21}.
Figure~\ref{fig:metric_fail_case} shows a specific case.

To alleviate the inaccuracy caused by the above reasons, we mainly use the more strict metrics (i.e., Leaves/Steps/Intermediates/Overall AllCorrect) for comparison.
Furthermore, we adopt manual evaluation on the full trees and individual steps to make a more accurate comparison (Sec.~\ref{sec:manual_eval}).

\section{Case Study}
\label{Sec:case_study}
\begin{figure*}[b]
\centering
\includegraphics[width=\textwidth]{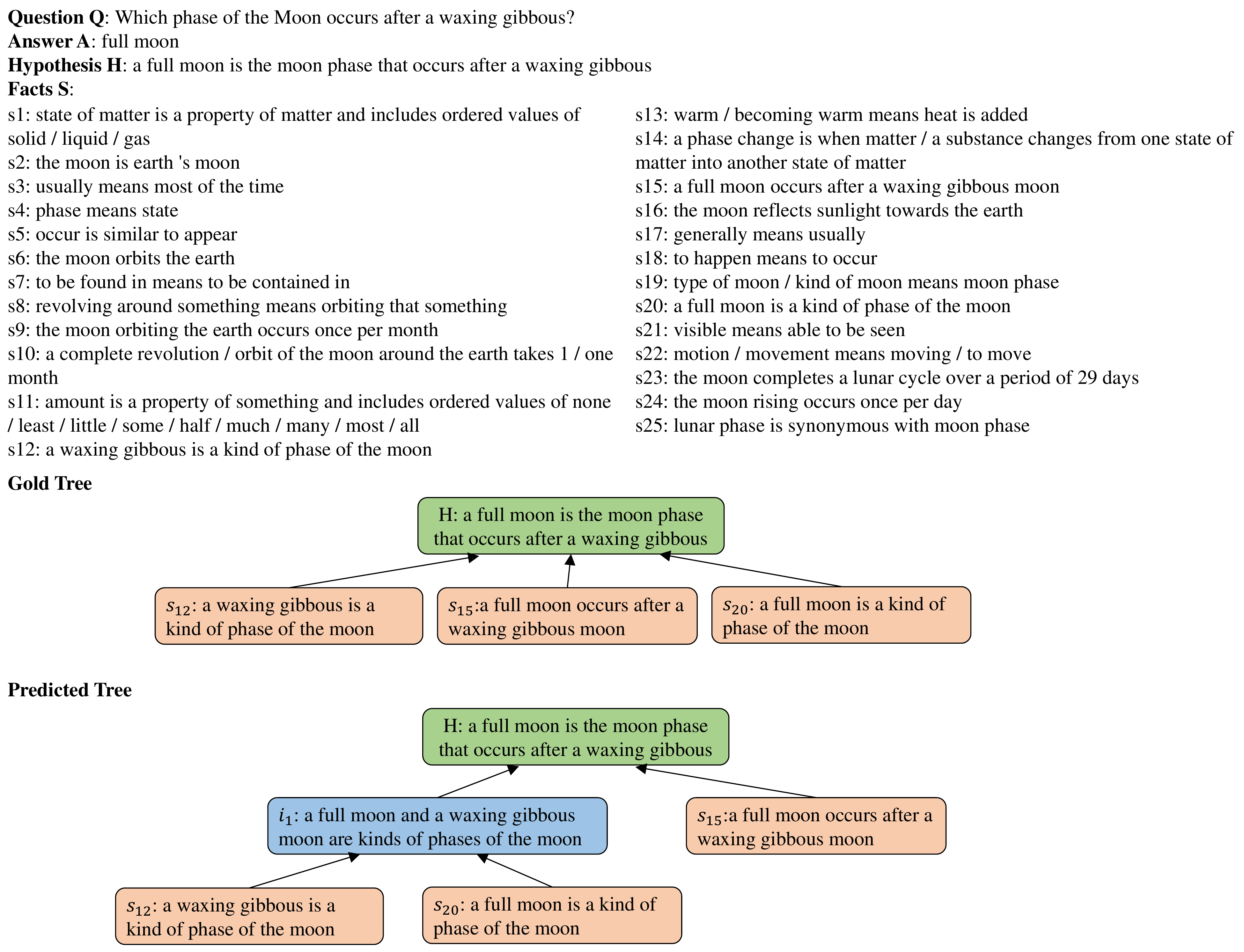}
\caption{Case 1. 
The predicted entailment tree consists of two 2-premise steps, while the gold tree consists of one 3-premise step.
Under the automatic evaluation metric, the predicted tree would be rated as invalid (Overall AllCorrect = 0), since the predicted steps do not match the gold step.
However, the predicted tree should be valid because each step in the tree is a valid entailment (i.e., the 3-premise step can be decomposed into two valid 2-premise steps).
It would be rated as valid under manual evaluation.
}
\label{fig:case1}
\end{figure*}

\begin{figure*}[t]
\centering
\includegraphics[width=\textwidth]{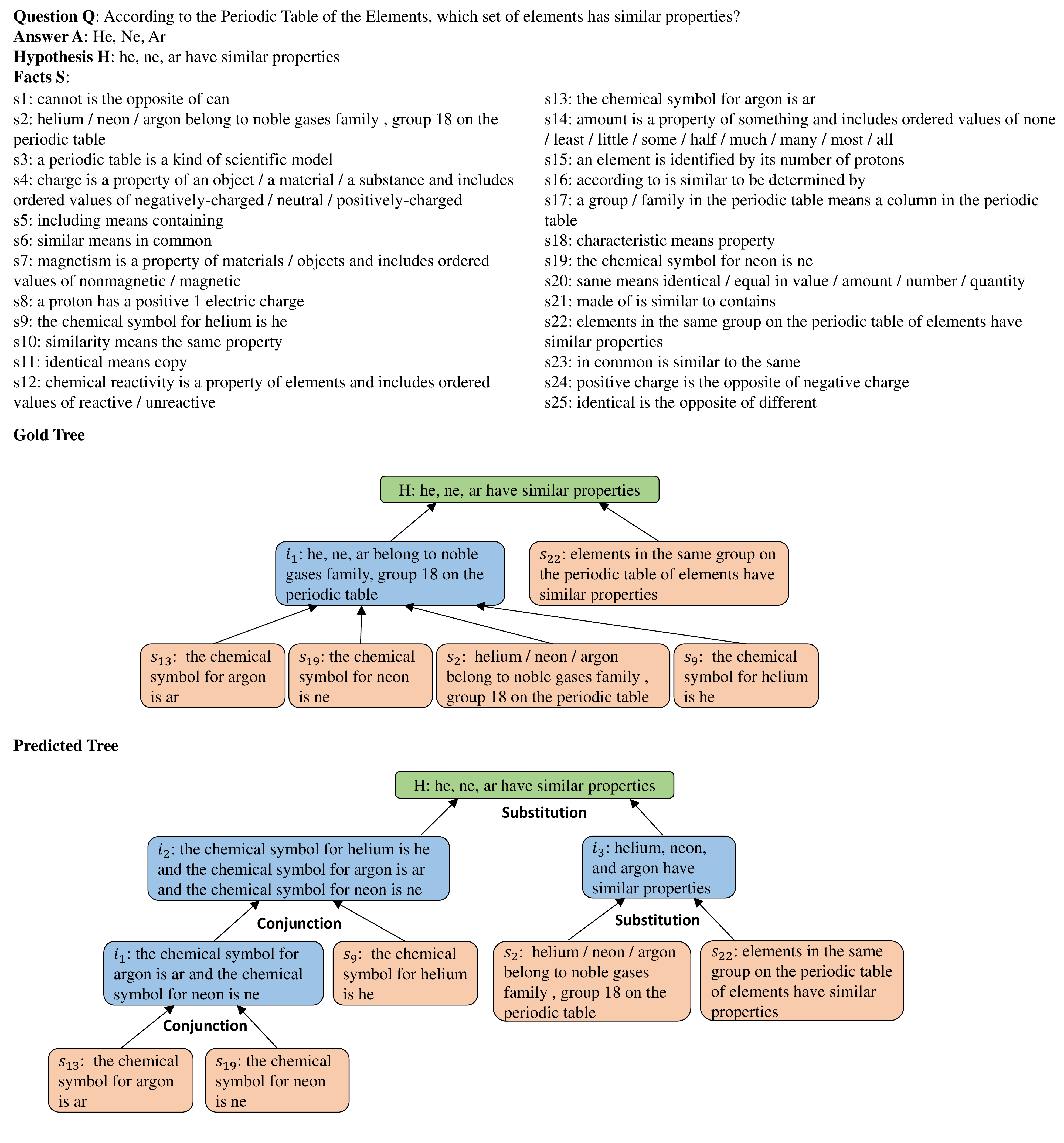}
\caption{Case 2.
Explaining the question and answer in this case requires 5 leaf facts from the given 25 facts.
\modelname can select the correct facts, generate valid entailment trees, and provide the reasoning types of steps.
}
\label{fig:case2}
\end{figure*}

\begin{figure*}[t]
\centering
\includegraphics[width=\textwidth]{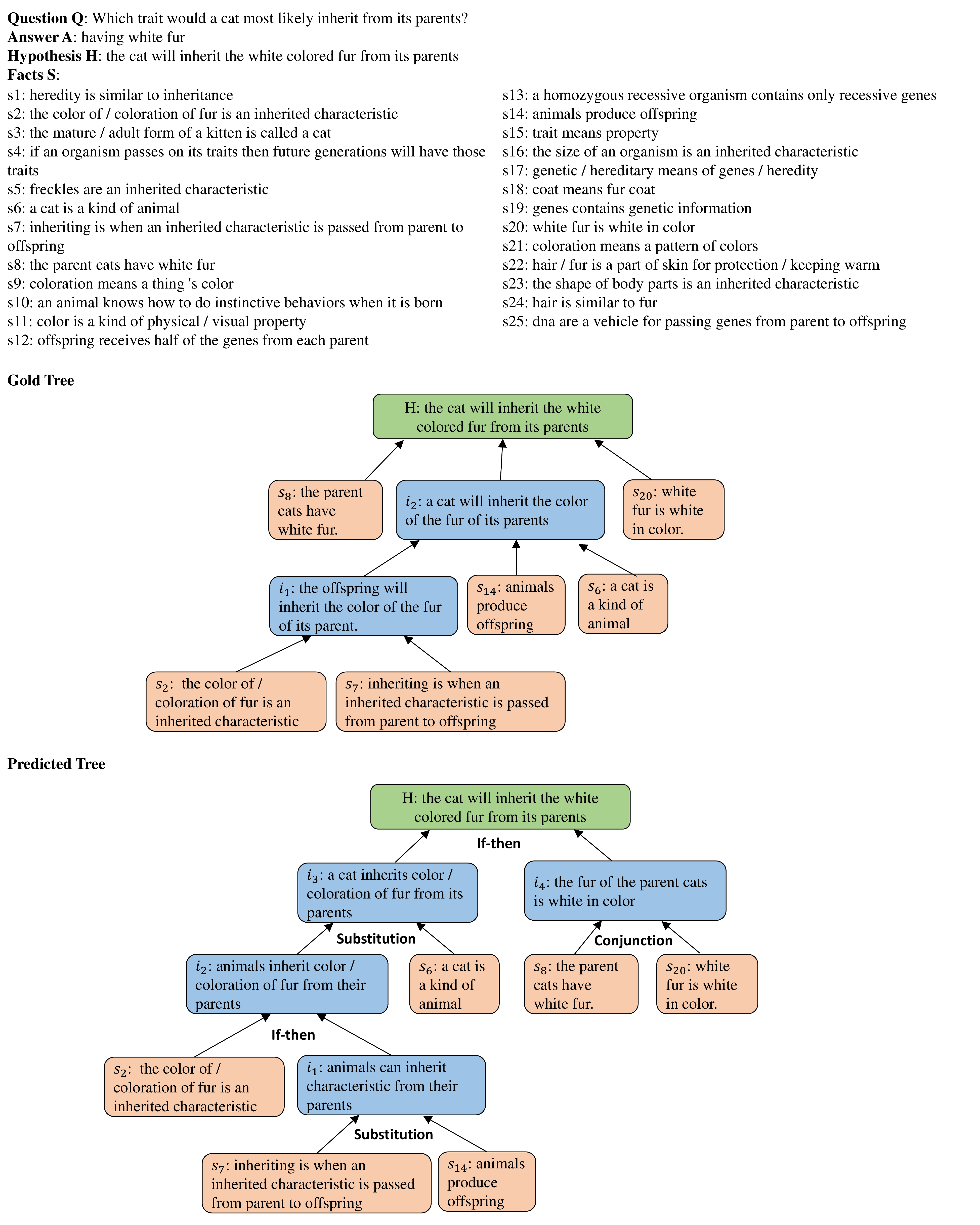}
\caption{Case 3.
\modelname can handle medium-complexity questions and provide the reasoning types of steps.
}
\label{fig:case3}
\end{figure*}

\begin{figure*}[t]
\centering
\includegraphics[width=0.95\textwidth]{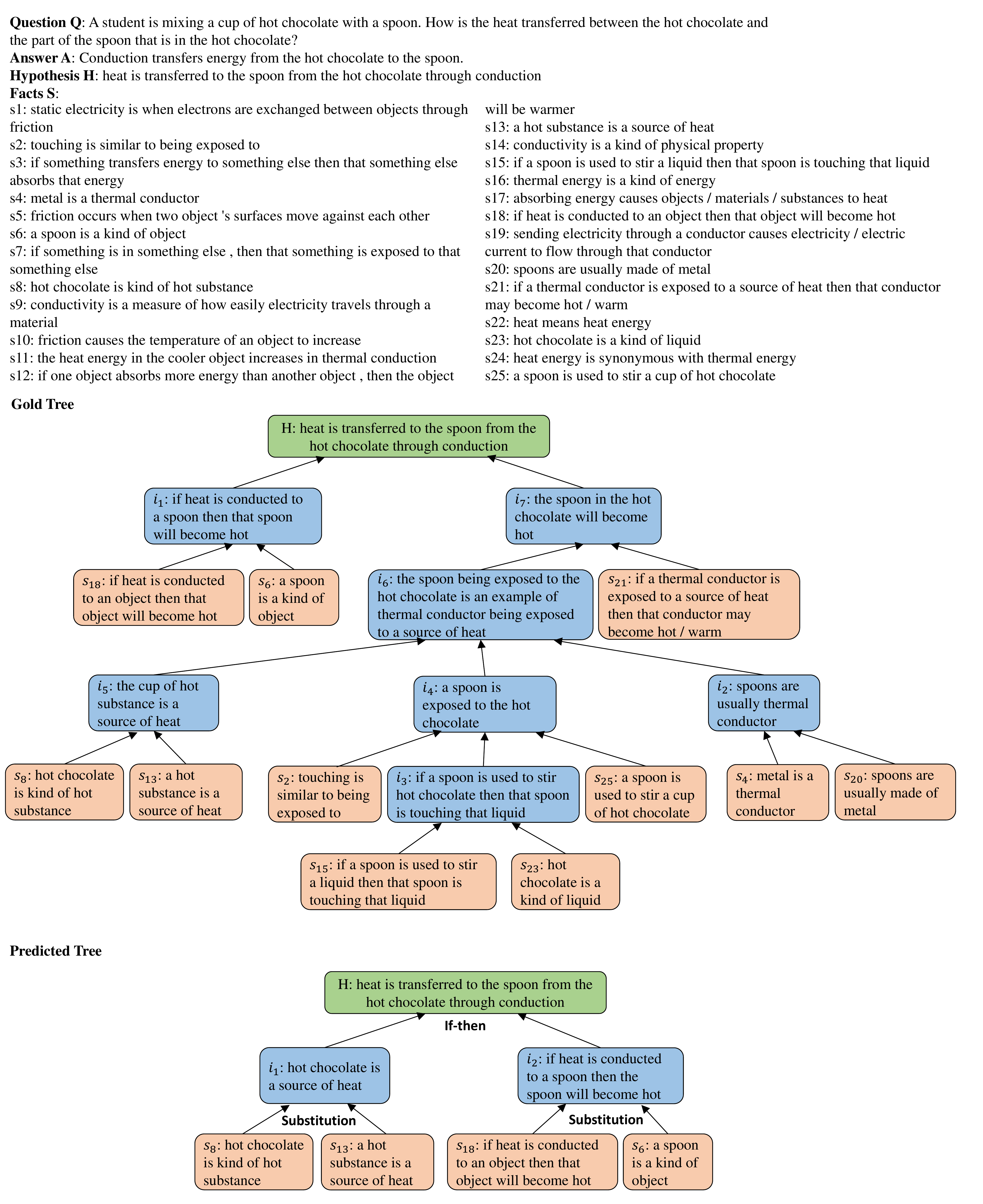}
\caption{Case 4.
The question requires more complex reasoning, where the gold tree contains 11 leaf facts and 8 entailment steps.
Although the full tree generated by \modelname is not entirely correct, the intermediate conclusions $i_1$, $i_2$ are still reliable.}
\label{fig:case4}
\end{figure*}

We show some entailment trees generated by our \Oursep on the Task2 questions in Figure~\ref{fig:case1},~\ref{fig:case2},~\ref{fig:case3},~\ref{fig:case4}.
\modelname can generate a valid entailment tree which may have a different structure with the gold one (Figure~\ref{fig:case1}).
\modelname can handle medium-complexity questions, generate valid entailment trees and provide the reasoning types of steps (Figure~\ref{fig:case2} and~\ref{fig:case3}).
The questions which require more complex reasoning (e.g., the gold tree in Figure~\ref{fig:case4} requires 11 leaf facts and 8 entailment steps) remain challenging.
Although the full tree generated by our method for such complex question can be not entirely correct, the intermediate conclusions (e.g., $i_1$, $i_2$ in Figure~\ref{fig:case4}) are still reliable.

\end{document}